\documentclass[lettersize,journal]{IEEEtran}
\usepackage{amsmath,amsfonts}
\usepackage{algorithmic}
\usepackage{algorithm}
\usepackage{array}
\usepackage[caption=false,font=normalsize,labelfont=sf,textfont=sf]{subfig}
\usepackage{textcomp}
\usepackage{stfloats}
\usepackage{url}
\usepackage{verbatim}
\usepackage{graphicx}
\usepackage{booktabs}
\usepackage{cite}
\usepackage{times}
\usepackage{epsfig}
\usepackage{amssymb}
\usepackage[accsupp]{axessibility}  
\usepackage{bbding}
\usepackage[pagebackref=true,breaklinks=true,letterpaper=true,colorlinks,bookmarks=false]{hyperref}
\newcommand{\etal}{\textit{et al}. }
\newcommand{\ie}{\textit{i}.\textit{e}. }

\usepackage{soul}

\begin{document}

\title{EvaSurf: Efficient View-Aware Implicit Textured Surface Reconstruction}

\author{Jingnan Gao, Zhuo Chen, Yichao Yan, Bowen Pan, Zhe Wang, Jiangjing Lyu, Xiaokang Yang$^*$,~\IEEEmembership{Fellow,~IEEE}
\thanks{($^*$Corresponding author: Xiaokang Yang.)}
\thanks{Jingnan Gao, Zhuo Chen, Yichao Yan and Xiaokang Yang are with the MoE Key Lab of Artificial Intelligence, AI Institute, Shanghai Jiao Tong University, Shanghai, China. (email: gjn0310@sjtu.edu.cn; ningci5252@sjtu.edu.cn; yanyichao@sjtu.edu.cn; xkyang@sjtu.edu.cn)}
\thanks{Bowen Pan, Zhe Wang and Jiangjing Lyu are with the Alibaba Group, Zhejiang, China. (email: bowen.pbw@alibaba-inc.com; xingmi.wz@alibaba-inc.com; jiangjing.ljj@alibaba-inc.com)}
\thanks{Project page: \url{https://g-1nonly.github.io/EvaSurf-Website/}.}
}

% The paper headers
\markboth{Submitted to IEEE Transactions on visualization and computer graphics}
{Shell \MakeLowercase{\textit{et al.}}: A Sample Article Using IEEEtran.cls for IEEE Journals}

\maketitle
\begin{figure*}[h]
    \centering
    \includegraphics[width=\linewidth]{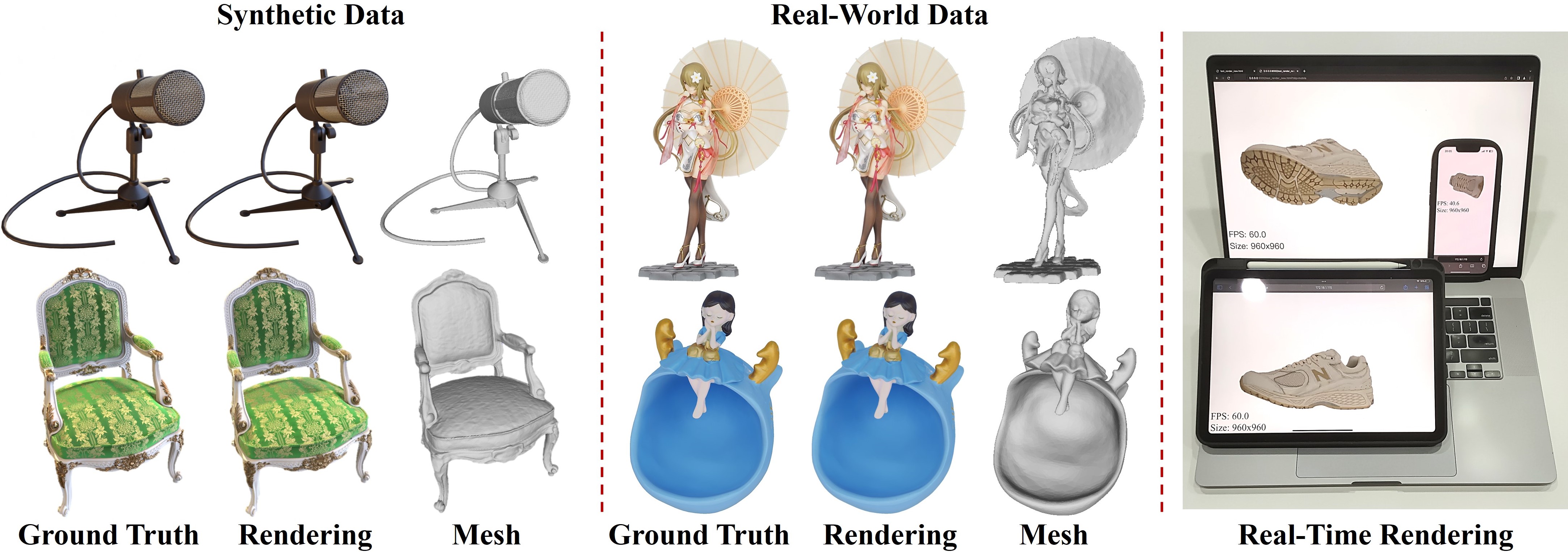}
    \caption{Examples of reconstruction results of \textbf{EvaSurf}. Our model can reconstruct high-quality appearance and accurate mesh for both synthetic and real-world objects. EvaSurf also supports real-time rendering on various devices.}
    \label{fig:teaser}
    \vspace{-5mm}
\end{figure*}

\begin{abstract}
Reconstructing real-world 3D objects has numerous applications in computer vision, such as virtual reality, video games, and animations. Ideally, 3D reconstruction methods should generate high-fidelity results with 3D consistency in real-time. Traditional methods match pixels between images using photo-consistency constraints or learned features, while differentiable rendering methods like Neural Radiance Fields (NeRF) use differentiable volume rendering or surface-based representation to generate high-fidelity scenes. However, these methods require excessive runtime for rendering, making them impractical for daily applications. To address these challenges, we present \textbf{EvaSurf}, an \textbf{E}fficient \textbf{V}iew-\textbf{A}ware implicit textured \textbf{Surf}ace reconstruction method on mobile devices. In our method, we first employ an efficient surface-based model with a multi-view supervision module to ensure accurate mesh reconstruction. To enable high-fidelity rendering, we learn an implicit texture embedded with view-aware encoding to capture view-dependent information. Furthermore, with the explicit geometry and the implicit texture, we can employ a lightweight neural shader to reduce the expense of computation and further support real-time rendering on common mobile devices. Extensive experiments demonstrate that our method can reconstruct high-quality appearance and accurate mesh on both synthetic and real-world datasets. Moreover, our method can be trained in just 1-2 hours using a single GPU and run on mobile devices at over 40 FPS (Frames Per Second), with a final package required for rendering taking up only 40-50 MB.
\end{abstract}


\section{Introduction}
\label{sec:intro}
3D reconstruction has long been an essential task in computer graphics and computer vision, with broad potentials in the area of VR/AR, video games, and film industry~\cite{AndersenVP19,HuangDGN17, LiuCKKH21}. 
To support the generalized applications, it is desirable for an ideal 3D reconstruction method to have the following  properties: 
(1) \textbf{High efficiency} that supports real-time rendering for 3D objects on various devices including PCs, tablets, and mobile phones; 
(2) \textbf{Realistic rendering} that faithfully reflects the appearance of original objects with multi-view consistency; 
(3) \textbf{Accurate geometry} that provides a detailed mesh to be flexibly manipulated by users with modern software. 

Recently, neural rendering methods~\cite{LombardiSSSLS19,LoubetHJ19,LuanZBD21,LyuWLC020, Martin-BruallaR21, MildenhallSTBRN20, Niemeyer021, NiemeyerMOG20, PumarolaCPM21, wang2021nerfmm, YuYTK21,DBLP:conf/cvpr/VerbinHMZBS22,nerv2021,DBLP:journals/tog/ZhangSDDFB21,Barron_2023_ICCV,9909994,peng2023intrinsicngp,10144678} have emerged as a promising solution because they achieve high-fidelity rendering results from multi-view images. 
However, traditional NeRF-based methods~\cite{Martin-BruallaR21, MildenhallSTBRN20,DBLP:conf/iccv/ParkSBBGSM21,DBLP:journals/corr/abs-2010-07492,wang2021nerfmm} require plenty of sampling points to forward a deep MLP network, requiring massive computation that is impractical for mobile devices. 
To improve the efficiency of neural rendering, current methods~\cite{SNeRG,mobilenerf} usually store the opacity and texture using representations that simplify the sampling process of the neural radiance field. 
While these methods achieve efficient and photo-realistic rendering on mobile devices, they suffer from the limitation in capturing accurate geometry as the continuous implicit function cannot provide explicit information about the underlying geometry of the object. 
Recent Gaussian-based methods~\cite{2dgs,3dgs,scaffoldgs,Yu2023MipSplatting} achieves high-quality rendering with high efficiency by optimizing an explicit representations. Nevertheless, due to the requirement of large disk storage and the lack of mobile-end optimization, they fail to render at interactive framerates on mobile devices.
To obtain accurate geometry, the following works~\cite{rerend, nerf2mesh, bakedsdf,DBLP:journals/tog/ReiserSVSMGBH23} for efficient rendering adopt surface-based methods that learn an implicit-explicit hybrid representation and then bake it into meshes.
Despite the improvement in geometry, they are still insufficient to obtain accurate meshes with smooth and solid surfaces regarding complex structures or non-Lambertian surfaces. 
Besides, due to the entangled learning of appearance and material, it is also difficult for these methods to render regions with view-dependent effects such as specularities. 

Based on the above discussions, it is natural to ask a question: can we build a model that solves the mentioned challenges and meanwhile satisfies all three requirements, \ie, high efficiency, realistic rendering, and accurate geometry? 
In this work, we present \textbf{EvaSurf}, an \textbf{E}fficient \textbf{V}iew-\textbf{A}ware implicit textured \textbf{Surf}ace reconstruction method on mobile devices. As the joint learning for the geometry and appearance may lead to entangled attributions, we design a two-stage framework to respectively reconstruct the geometry and appearance. 
\textbf{(1)} In the first stage, we employ an efficient surface-based model with a multi-view supervision module to learn accurate structures of geometry.
\textbf{(2)} In the stage of appearance reconstruction, the view-dependent effect severely influences the quality of rendering images. Therefore, we incorporate view-aware encoding into an implicit texture, to assign different weights to the dimensions of the view-aware texture, thus modeling the appearance with view-dependent effects.
\textbf{(3)} With the explicit geometry and the implicit texture, we show that a lightweight neural shader is sufficient to achieve high-quality differentiable rendering. The small size of the neural shader can greatly reduce the consumption of computation and memory sources, empowering real-time rendering on mobile devices. 

With the help of proposed strategies, the training procedure of our method can be completed in just 1-2 hours on a single GPU, 10$\times$ faster than MobileNeRF~\cite{mobilenerf}. 
The inference rendering can be performed on common mobile devices at a speed of at least 40 FPS (Frames Per Second), and the final storage of rendering packages only takes up 40-50 MB (Megabytes), much smaller than existing methods~\cite{mobilenerf, rerend, nerf2mesh}.
Moreover, as depicted in Fig.~\ref{fig:teaser}, our method can not only handle synthetic data but also reconstruct high-quality results on complex real-world objects. 

In summary, our framework, EvaSurf:
\begin{itemize}
    \item proposes a two-step framework to achieve efficient, high-quality 3D reconstruction while obtaining accurate mesh;
     \item models the view-dependent appearance by incorporating the view-aware encoding into an implicit texture;
    \item runs on common devices including PCs, tablets, and mobile phones with over 40 FPS and trains in 1-2 hours while consuming much smaller disk storage of the rendering package than existing methods.
    
\end{itemize}

\section{Related Work}
\label{sec:relatedwork}
\subsection{Neural Rendering for 3D Reconstruction}
To ensure the consistency of reconstructed 3D objects, multi-view methods~\cite{Blumenthal-BarbyE14,CampbellVHC08,GallianiLS15, TolaSF12, VuLPK12,ChenLGSLJF19,Liu0LL19} use geometry representations like point clouds and depth maps to extract the corresponding mesh. Later methods~\cite{EstebanS03,FuaL95,LiaoDG18,MunkbergCHES0GF22,ShenGYLF21,DBLP:conf/cvpr/JiangSMHNF20} estimate the mesh by optimizing an initial mesh based on photo consistency. Furthermore, learned features and neural shape priors were introduced~\cite{LinWRSKFL19, WenZCLXF23,WenZLF19} for mesh deformation. These methods often rely on strict assumptions to ensure 3D consistency, which cannot yield high-fidelity results for varying light conditions. 

More recently, neural rendering technique has been widely adopted in the field of 3D object reconstruction. 
Incorporating the implicit representations, Neural Radiance Field (NeRF)~\cite{MildenhallSTBRN20} and its followers~\cite{Martin-BruallaR21,Niemeyer021,PumarolaCPM21,wang2021nerfmm,mipnerf,DBLP:conf/cvpr/BarronMVSH22,DBLP:conf/iccv/ReiserPL021,DBLP:conf/iccv/GarbinK0SV21,DBLP:journals/tog/MullerESK22,DBLP:conf/eccv/ChenXGYS22,DBLP:conf/cvpr/Fridovich-KeilM23} achieve high-fidelity rendering results. These methods incorporate ray-marching and ray-sampling mechanisms during the process of neural rendering through an encoded 5D continuous field. The neural network estimates volume densities and RGB values along the sampled points on the marching rays.
However, the corresponding geometry reconstructed by these methods is not satisfactory and accurate due to the ambiguity of volume rendering. 
Besides implicit representation, recent 3D Gaussian splatting~\cite{3dgs, 2dgs,scaffoldgs,Yu2023MipSplatting, r3dgs} involves iterative refinement of Gaussians to reconstruct 3D objects from multiview images, allowing for high-quality rendering of novel views. Although Gaussian-based methods can achieve real-time rendering on PCs, real-time rendering on mobile devices remains a challenge due to the requirement of large disk storage and lack of mobile-end neural network optimization.

To improve the quality of geometry, surface-based methods open up another road for 3D reconstruction. These methods optimize the underlying surface directly, or gradually deform the geometry from volumetric representation into a surface representation. The implicit surface representations are encoded in neural networks such as occupancy networks~\cite{MeschederONNG19,NiemeyerMOG20,OechsleP021,DBLP:conf/eccv/PengNMP020} or neural signed distance functions (SDF)~\cite{NiemeyerMOG20,OechsleP021,ViciniSJ22,neus,YarivKMGABL20,DBLP:conf/nips/Fu0OT22,DBLP:conf/iclr/WuWPXTLL23,DBLP:conf/cvpr/Li0ETU0L23,DBLP:conf/iccv/YuLT0NK21,DBLP:conf/cvpr/Fridovich-KeilY22,DBLP:conf/nips/YarivGKL21,DBLP:conf/nips/YuPNS022,bakedsdf,neudf}. 
These SDF-based models can reconstruct a solid mesh with details. However, such methods tend to generate a mesh that lacks smoothness and suffers from geometric ambiguities, resulting in inaccurate shape. Particularly, concave surfaces are easily interpreted as convex oversmooth surface using such methods.In addition, these methods often require excessive runtime and thus become impractical for daily real-time applications. Multi-view stereo (MVS) reconstruction techniques are then adapted to NeRF~\cite{d1nerf,d2nerf,nerfingmvs} for better shape reconstruction. These methods can reconstruct a smoother mesh with an accurate representation of convex and concave shapes. Nevertheless, such methods are prone to generate a damaged and incomplete mesh. Conversely, our method adopts neural rendering for the appearance but uses explicit mesh to represent the geometry. We implement our method into an efficient two-stage framework and supports real-time rendering. Furthermore, we employ an additional supervision module to take advantage of both SDF and MVS methods for better reconstruction.

\subsection{Real-time Rendering on Mobile Devices}
With the rapid development of modern technology, mobile phones, and other mobile devices have become an essential part of people's daily lives. Real-time rendering on these devices has emerged as an important research topic. Before the introduction of neural radiance fields, several methods~\cite{DBLP:journals/tog/ThiesZN19,Wang_2021_CVPR,DBLP:conf/iclr/ThiesZTSN20,nerftex} employ a neural texture and a neural renderer to render a scene. Though these methods can achieve photo-realistic rendering in real time, they may not be generalized to arbitrary 3D objects due to their requirements of geometry priors.

To facilitate the power of NeRF to render scenes on mobile devices, Hedman \etal~\cite{SNeRG} propose a solution that bakes NeRF into sparse neural radiance grids (SNeRG). However, the rendering speed of SNeRG fails to achieve real-time performance, hindering the attainment of interactive rendering rates. In contrast, MobileNeRF~\cite{mobilenerf} adopts textured polygons as the scene representation and leverages the polygon rasterization pipeline to accelerate the rendering process. 
Although MobileNeRF supports real-time rendering with interactive frame rates, the exported mesh may be inaccurate due to its complex interlacing polygons. 
Re-ReND~\cite{rerend} and NeRF2Mesh~\cite{nerf2mesh} both train a NeRF-like model at the initial stage and then bake the appearance into a texture map and extract the corresponding mesh. Nevertheless, these methods~\cite{rerend,nerf2mesh,rakotosaona2023nerfmeshing} are unable to render scenes with view-dependent effects and can not reconstruct accurate mesh for complex structures and non-Lambertian objects. Bakedsdf~\cite{bakedsdf} utilizes a hybrid SDF-volume representation and bakes the representation into meshes for real-time rendering. However, it tends to learn over-smooth surfaces and cannot handle complex structures. In comparison, our method employs a multi-view supervision module to learn the complex structure and prevent the model from over-smoothing. We also utilize a view-aware encoding in feature space for better renderings.

\section{Methods}
\label{sec:methods}
\begin{figure*}[tb]
    \centering
    \includegraphics[width=\linewidth]{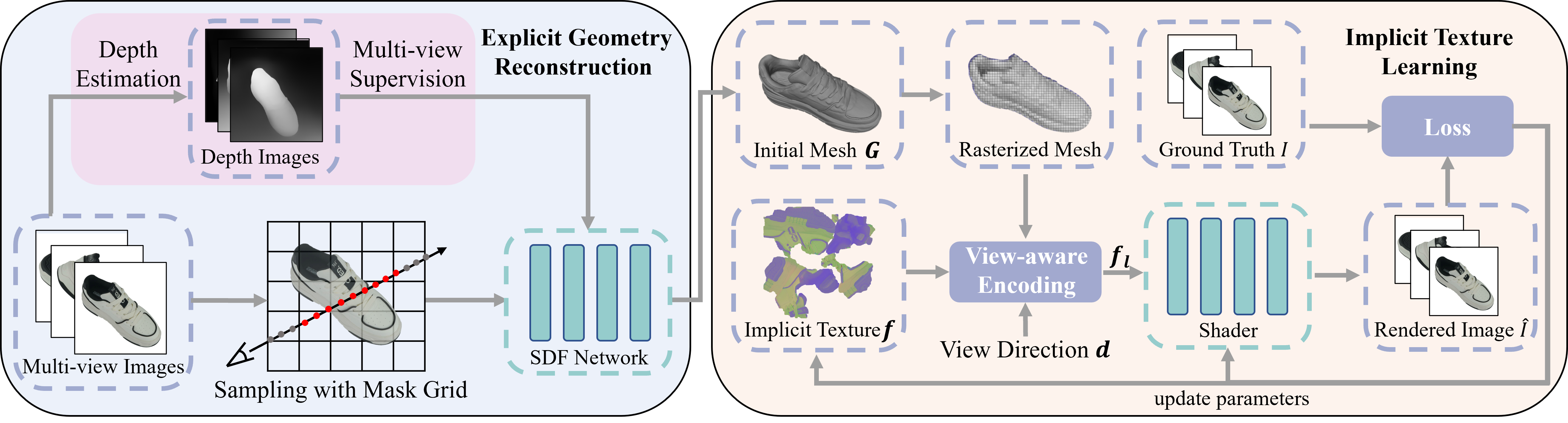}
    \caption{\textbf{The overview of our full pipeline}. We begin with an initial explicit geometry generated by a surface-based model. We then rasterize the mesh and utilize view-aware encoding to embed the view-dependent information into a learnable implicit texture. Given the geometry and texture, a neural shader renders the final RGB images.}
    \label{pipeline}
\end{figure*}
Given a set of images $\mathcal{I}=\{I_{1}, ..., I_{n}\}$, we aim to optimize a representation to efficiently generate accurate mesh and support synthesizing novel views at interactive frame rates.
To this end, we design a two-stage differentiable framework that requires much less computation and disk storage so as to make it compatible with mobile devices.
As shown in Fig.~\ref{pipeline}, we first utilize a surface-based model $\mathcal{S}$ to reconstruct an accurate geometry (\ref{geo_init}).
In the second stage, we rasterize the initial mesh and introduce a view-aware encoding $L$ to embed the view-dependent information into an implicit texture $f$ (\ref{lighting}).
Finally, we jointly optimize the implicit texture $f$ and a neural shader, \ie, a lightweight MLP, which takes the view-embedded implicit feature $f_l$ as input and directly outputs the rendered images $\hat{I}$ (\ref{shader}).
\subsection{Efficient Explicit Geometry Reconstruction}
\label{geo_init}
As the base of the reconstruction pipeline, an accurate geometry is crucial for the high-quality rendering.
To overcome the challenge of coarse geometry faced by the traditional NeRFs, our model first trains a surface-based model $\mathcal{S}$ that follows the pipeline of NeuS~\cite{neus} and leverages a set of signed distance functions to directly estimate the explicit geometry. 

However, previous surface-based models usually require excessive training time and struggle to generate accurate results for objects with complex structures.
Therefore, we design two additional strategies to further solve these challenges. To tackle the issue of low-speed training, as depicted in Fig.~\ref{pipeline}, we employ a set of spatial grids with mask to filter the sampled points during the learning process. We update the mask grid for each training iteration by recording the points with SDF values less than zero. By caching this mask grid to skip known freespace during ray-sampling, we enable the model to converge within 2 hours (\textbf{over 10$\times$ faster than Neuralangelo~\cite{DBLP:conf/cvpr/Li0ETU0L23}}) through dynamic sampling. It is noteworthy that employing spatial grids at low resolution may lead to smooth but coarse mesh results, while high-resolution grids might induce overfitting. In order to generate an accurate geometry without introducing additional noise, we utilize a set of progressive grids $g(\beta)$ that gradually increase the bandwidth to realize the coarse-to-fine training strategy. The set of progressive grids can be formulated as follows:
\begin{equation}
    g(\beta) = (w_1(\beta)g_1,w_2(\beta)g_2, ... , w_l(\beta)g_l),
\end{equation}
where $\beta$ is a parameter determined empirically to modulate the bandwidth and $g_i$ is the multi-resolution spatial grids at level-$i$ with corresponding weight indicator $w = I(i\leq\beta)$.

To enable the surface-based model to learn the accurate shape of complex structures, we introduce a multi-view supervision (MVS) module to supervise the SDF-based geometry learning. The MVS module first reconstructs the shape using the open multi-view stereo reconstruction library OpenMVS~\cite{openmvs} and renders depth images. We then supervise the surface-based model using the rendered depth images. The MVS module can prevent our model from generating over-smooth surfaces and encourage it to learn complex structures. 
Overall, for an explicit geometry $G$ represented by triangle mesh with vertex positions $V$, the objective is:
\begin{equation}
     \underset {V,\theta} { \operatorname {arg\,min} } \, (\mathcal{L}_s (G) + \mathcal{L}_{mvs}(G)),
\end{equation}
where $\theta$ is the parameters of the geometry network to be optimized, $\mathcal{L}_s$ is the loss of the surface-based model and $\mathcal{L}_{mvs}$ is the loss of the multi-view supervision module. 
Specifically, the formulation of $\mathcal{L}_s$ is demonstrated as follows:
\begin{equation}
    \begin{aligned}
    \mathcal{L}_s  =  w_{rgb} L_{rgb} + w_{eikonal} L_{eikonal} + w_{mask} L_{mask}, 
\end{aligned}
\end{equation}
where $L_{rgb}$ is the \textit{L2} loss, $L_{eikonal}$ is the loss term proposed in NeuS~\cite{neus} and $L_{mask}$ also follows the definition of NeuS~\cite{neus}. The weight for each loss is determined by default as $w_{rgb}=1.0$, $w_{eikonal}=0.1$, $w_{mask}=0.1$ in experiments.

 In the MVS module, we employ OpenMVS~\cite{openmvs} to reconstruct the shape and render depth images. We then supervise the surface-based model using the rendered depth images and the normals calculated from the depths. By employing this additional module, we introduce the depth priors to the surface-based model, encouraging the model to capture accurate shape information while reconstructing a solid and smooth surface. The effectiveness of this module is demonstrated in the experiments section of the paper. The formulation of $\mathcal{L}_{mvs}$ is demonstrated as follows:
\begin{equation}
    \mathcal{L}_{mvs} = w_{depth} L_{depth} + w_{normal} L_{normal},
\end{equation}
where $L_{depth}$ is the \textit{L1} loss between the depth generated by OpenMVS and SDF-model, and $L_{normal}$ is the \textit{L2} loss between the normal derived from the OpenMVS depths and the normal generated by the SDF-mode. The weight is determined by default as $w_{depth}=0.3$, $w_{normal}=0.04$  in experiments.

In contrast to the mesh generated by MobileNeRF~\cite{mobilenerf}, which is produced through the intersection of interlaced polygons, our mesh can be directly unwrapped into UV form and can be edited directly without further operation in contemporary software packages. Compared to previous methods~\cite{rerend,nerf2mesh} that extract a mesh from a trained NeRF model, our method can generate an explicit mesh with clearly defined boundaries and high accuracy. The normal information obtained from the explicit mesh is utilized for subsequent improvement in the texture training stage, establishing a robust foundation for the overall pipeline.

\subsection{View-aware Implicit Texture}
\label{lighting}
\noindent\textbf{Implicit Texture.}
Implicit representations like neural radiance fields or neural textures are usually introduced to capture high-frequency details, but they usually lead to large MLPs that requires massive computation. To improve the efficiency, current frameworks~\cite{nerf2mesh,rerend} directly bake the appearance into explicit texture maps with topological structures. Encouraged by these methods, we incorporate topological information into an implicit texture to improve shading efficiency while facilitating the capture of high-frequency details. 

Given an initialized geometry $G$, we first store the UV coordinates $(u,v)$ through mesh rasterization to build up a topological structure. Then we initialize a learnable feature map with random noise as the implicit texture and optimize it under the prior of the established UVs. 
This allows for the refinement of appearances with topological structure priors. On one hand, by introducing topological structures, the implicit texture can be directly mapped to the UVs, which reduces memory consumption of the neural shader. On the other hand, the potential to capture additional information arises from implicit representation compared with explicit texture.

\noindent\textbf{View-aware Encoding.}
Intuitively, a point is expected to exhibit varying colors when observed from different view directions, namely view-dependent effects. While the implicit texture can render photo-realistic scenes on mobile devices, it still struggles to model such effects, similar to current frameworks~\cite{nerf2mesh,rerend}. 
To capture view-dependent appearance, we then perform view-aware encoding to introduce additional view-dependent information to the implicit texture.

\begin{figure*}[tb]
    \centering
    \includegraphics[width=\linewidth]{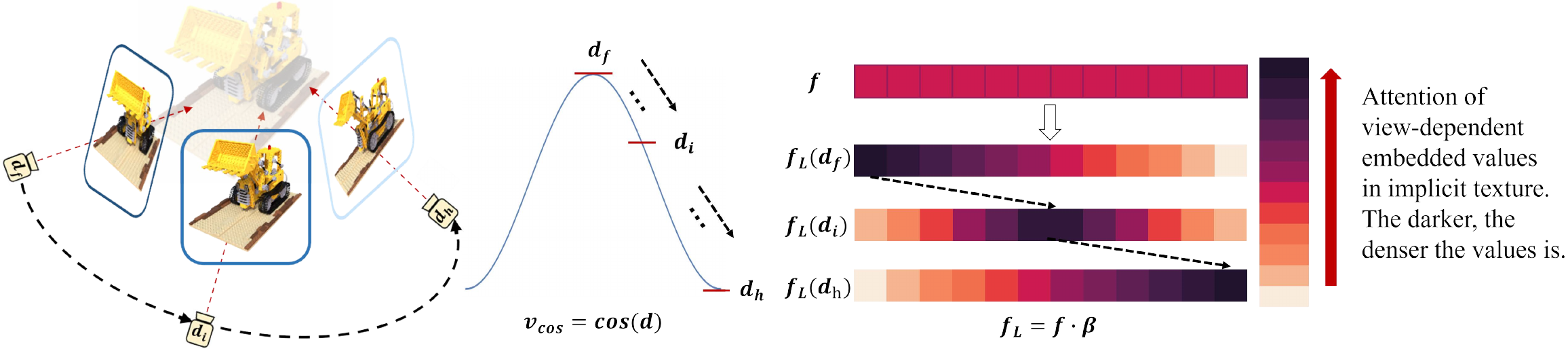}
    \caption{View-dependent embedded mechanism in feature space. We equip our model with a set of Gaussians to capture the view-dependent information. The heatmap demonstrates the corresponding relation between the values and the view direction.
    }
    \label{fv}
\end{figure*}

Specifically, we utilize Gaussians to assign different weights to the texture feature for various views, as depicted in Fig.~\ref{fv}. However, applying Gaussians at the stage of ray sampling or positional encoding usually requires additional computation, preventing our model from efficient rendering on mobile devices.
Therefore, unlike Mip-NeRF~\cite{mipnerf} that employs Gaussians to approximate the conical frustums on the 3D point level, our method instead utilizes the set of \textbf{Gaussians on the feature-level}. Furthermore, different from BakedSDF~\cite{bakedsdf} that embeds the spherical Gaussians with each vertex of the mesh, we adopt the traditional Gaussian that are distributed in the feature space. 
Specifically, given a view direction $d \in \mathbb{R}^3 $, we first calculate the normals $n \in \mathbb{R}^3$ at each position and then calculate a cosine value $v_{cos}$ that represents the angle between the view direction $d$ and the normals $n$, serving for the following view-aware encoding computation:
\begin{equation}
    v_{cos} = \frac{-d \cdot n}{\left \| d \right \| \cdot\left \| n \right \|} .
\end{equation}
Furthermore, we employ a set of Gaussians to incorporate the view-dependent information into the implicit texture:
\begin{equation}
    L(f, v_{cos}) = f \cdot (\frac{1}{\alpha\sqrt{2\pi}} e^{-\frac{(v_{cos}-t)^2}{2\alpha^2}}),
    \label{feature}
\end{equation}
where $f$ is the learnable implicit texture, $\alpha$ is a hyper-parameter and $t$ is a constant array to indicate the channel number. 

Specifically, the view-dependent information is represented as the concentration of the values in the texture map. By assigning different weights to each dimension(Chanel) of the texture, we model view dependence with different fidelity and thus encourage the texture to learn the view-dependent information in the scene. Furthermore, our method introduces the view-dependent information in the feature space instead of the 3D point space. Learning the additional view-independent information upon the topological-structured UVs, we captured the anisotropy while preserving the inference efficiency.

\subsection{Light-weight Neural Shader}
\label{shader}
In our two-stage framework mentioned above, we utilize an efficient surface-based model to learn the explicit geometry and employ an implicit texture to model the appearance. With the explicit geometry and the implicit texture, the neural shader can be extremely light.
This is because its function is just to connect the view-aware texture with the RGB values, instead of computing the features representing the geometry and fitting the RGB values from the inputs, as the large MLP used in NeRF~\cite{MildenhallSTBRN20}. Therefore, our framework utilizes a much smaller MLP than the original NeRF model, which significantly reduces the computation expense and storage resources, enabling real-time rendering on mobile devices. 
The overall shading process can be summarized as follows. Given the view-embeded implicit texture $f_l = L(f, v_{cos})$, where $v_{cos}$ represents the cosine value, $L$ represents the view-aware encoding, and $f$ represents the implicit texture, our framework optimizes a neural shader $Shader_{\theta}$, which is a lightweight MLP. The pipeline can be formulated as:
\begin{equation}
     \hat{C}= Shader_{\theta}(f_l),
\end{equation}
where $\hat{C}$ represents the RGB colors of the rendered images $\hat{I}$.

\subsection{Differentiable Training}
\label{dr}
We adopt the differentiable rendering technique for the training procedure. As depicted in Fig.~\ref{pipeline}, during the forward-pass procedure, our method renders a 2D image given the explicit geometry and the learnable implicit texture, which altogether affect the shape and appearance of a 3D object. As for the backward-pass stage, the gradient of the loss with respect to the input shape and appearance factors is computed given the gradient of the loss function defined on the output image pixels. Here, we use the \textit{L2} loss as the reconstruction loss:
\begin{equation}
    \mathcal{L} = || \hat{C} - C_{gt} ||^2_2 ,
\end{equation}
where $\hat{C}$ represents the predicted RGB value and $C_{gt}$ represents the ground truth value. Therefore, the derivation of the backward-pass stage is formulated as:
\begin{equation}
\frac{\partial \mathcal{L}}{\partial\left\{x, y, z, w\right\}}=\frac{\partial \mathcal{L}}{\partial\{u, v\}} \cdot \frac{\partial\{u, v\}}{\partial\left\{x, y, z, w\right\}} ,
    \label{backward}
\end{equation}
where $(x,y,z,w)$ is the homogeneous coordinates input from the world space and $(u,v)$ is the barycentric coordinates from the image space. Similar to the rendering pipeline of nvdiffrast~\cite{nvdiffrast}, the mesh is rasterized and the 3D positions are interpolated pixel-wisely in image space. Since we build a topological-structured UVs after rasterization, the appearance from stage 1 can be inherited to stage 2. This approach reduces the required training time for stage 2 by eliminating the need to learn the appearance from scratch.

\section{Experiments}
\label{sec:experiments}

\subsection{Experiment Setups}

\noindent\textbf{Synthetic Dataset.} We evaluate the performance on the 8 synthetic $360^{\circ}$ scenes from NeRF~\cite{MildenhallSTBRN20}.
It consists of 8 path-traced scenes with 100 images for training and 200 images for testing per scene. 

\noindent\textbf{Real-world Dataset.} To demonstrate the effectiveness of our method, we also construct a set of real-world high-resolution (\textbf{2K}) data. This dataset comprises \textbf{15 objects}, with each object consisting of over 200 images. These images are captured by \textbf{accurate camera poses}, facilitating precise correlation between the data and the corresponding imaging parameters. Specifically, we provide 3 types of camera poses for each object following the convention of Colmap~\cite{schoenberger2016mvs,schoenberger2016sfm}, NeRF (\textit{transforms.json})~\cite{MildenhallSTBRN20} and NeuS (\textit{cameras\_sphere.npz})~\cite{neus}. The corresponding masks for the images are also included in the dataset.

\noindent\textbf{Network Architecture.}
For the network architecture of our surface-based model, we first use TCNNHashGrid from~\cite{DBLP:journals/tog/MullerESK22} with progressive levels to embed the points in 3d space. In the hierarchy of the grids, we use 16 levels from 16 to 4096, where each levels has 4 channels. The SDF MLP in our model has 2 layers with 64 hidden units. As for the texture network, it is noteworthy that unlike other SDF-based method the texture network in our surface-based model mainly serves to aid the geometry learning instead of producing the final rendering. Learning geometry from multi-view inputs can be a challenging task, so we adopt multi-head texture network during the pipeline. To be more specific, we adopt two texture networks during the geometry learning. We use positional encoding to embed the positions and spherical harmonics encoding to embed the view directions in the first texture network. In the second network, we use positional encoding for both the positions and view-directions. The texture network with multi texture heads accomplishes the task of coarse-to-fine learning. The first texture MLP has 2 layers with 64 hidden units and the second has 3 layers with 64 hidden units.

\begin{table*}[tb]
    \centering
    \caption{Comparison of Chamfer Distance $\downarrow$ on NeRF synthetic dataset (the unit is $10^{-3}$). We use ``Mobile" to denote whether a method supports rendering on mobile devices or not. Our method achieves better surface reconstruction quality compared to previous methods while supporting real-time rendering on mobile devices. Note that Re-ReND~\cite{rerend} extracts a mesh before the rendering pipeline but some of the meshes are not provided, so we only evaluated the meshes provided by the authors.}
    \begin{tabular}{l|c|cccccccc|c}
    \hline  & Mobile & Chair & Drums & Ficus & Hotdog & Lego & Materials & Mic & Ship & Mean \\
    \hline NeuS~\cite{neus}& \XSolidBrush & \textbf{3.95} & 6.68 & 2.84 & 8.36 & 6.62 & 4.10 & 2.99 & 9.54 & 5.64 \\
    NVdiffrec~\cite{MunkbergCHES0GF22}& \XSolidBrush & 4.13 & 8.27 & 5.47 & 7.31 & 5.78 & 4.98 & 3.38 & 25.89 & 8.15 \\
    Neuralangelo~\cite{DBLP:conf/cvpr/Li0ETU0L23}& \XSolidBrush              & 14.50 & 16.99 & 5.72  & 14.27    & 6.90 & 3.27 & 8.78 & 16.02 & 10.81 \\ 
     RelightableGaussian~\cite{r3dgs} & \XSolidBrush & 4.06 & 7.63 & 3.78 & 5.42 & 5.54 & 3.75 & 6.18 & 8.76 & 5.63 \\
     2DGS~\cite{2dgs}& \XSolidBrush & 5.25 & 10.33 & 4.41 & 9.55 & 6.74 & 9.09 & 11.06 & 9.55 & 8.25\\
    \hline MobileNeRF~\cite{mobilenerf}& \Checkmark & 7.82 &  9.49 & 8.98  & 11.23 &  9.96 & 8.31 & 9.61 & 15.23 & 10.08  \\
    Re-ReND~\cite{rerend}& \Checkmark& 3.96 & - & 5.60 & 11.04 & \textbf{4.65} & 3.40 & - & - & - \\
    BakedSDF~\cite{bakedsdf} & \Checkmark             & 4.05 & 7.41 & 3.23 & 6.72    & 5.69 & 5.39 & 3.17 & 8.98 & 5.58 \\
    NeRF2Mesh~\cite{nerf2mesh}&\Checkmark & 4.60 & \textbf{6.02} & \textbf{2.44} & 5.19 & 5.85 & 4.51 & 3.47 & 8.39 & 5.06 \\
    \hline Ours & \Checkmark & 4.16 & 6.97 & 2.72 & \textbf{4.48} & 5.33 & \textbf{3.23} & \textbf{2.83} & \textbf{5.42} & \textbf{4.39} \\
    \hline
    \end{tabular}
    \label{mesh_results}
\end{table*}

\subsection{Comparisons on Synthetic Data}
\noindent\textbf{Geometry Quality.} To evaluate the geometry quality, we compare the exported meshes of our method and previous methods~\cite{neus,MunkbergCHES0GF22,mobilenerf,rerend,nerf2mesh,2dgs,bakedsdf,DBLP:conf/cvpr/Li0ETU0L23}. We choose bi-directional Chamfer Distance as the evaluation metric. Nevertheless, due to the fact that the ground truth meshes may not accurately represent surface meshes, we cast rays and sample 2.5 million points from the intersections between the rays and the surfaces in each scene following the evaluation process of~\cite{nerf2mesh}. The quantitative results are demonstrated in Table~\ref{mesh_results}. Compared with previous \textbf{implicit} methods for surface reconstruction~\cite{neus,MunkbergCHES0GF22,DBLP:conf/cvpr/Li0ETU0L23}, our method achieves superior mesh quality while supporting real-time rendering on mobile devices. Compared with state-of-the-art \textbf{explicit} Gaussian-based methods~\cite{2dgs,r3dgs}, EvaSurf also reconstructs more accurate geometry for all the scenes while achieve mobile-rendering. Compared with existing methods~\cite{mobilenerf,rerend,nerf2mesh,bakedsdf} that allow rendering on mobile devices, our method can generate more accurate mesh for the scenes with view-dependent effects. 

\begin{figure*}[tb]
    \centering
    \includegraphics[width=\linewidth]{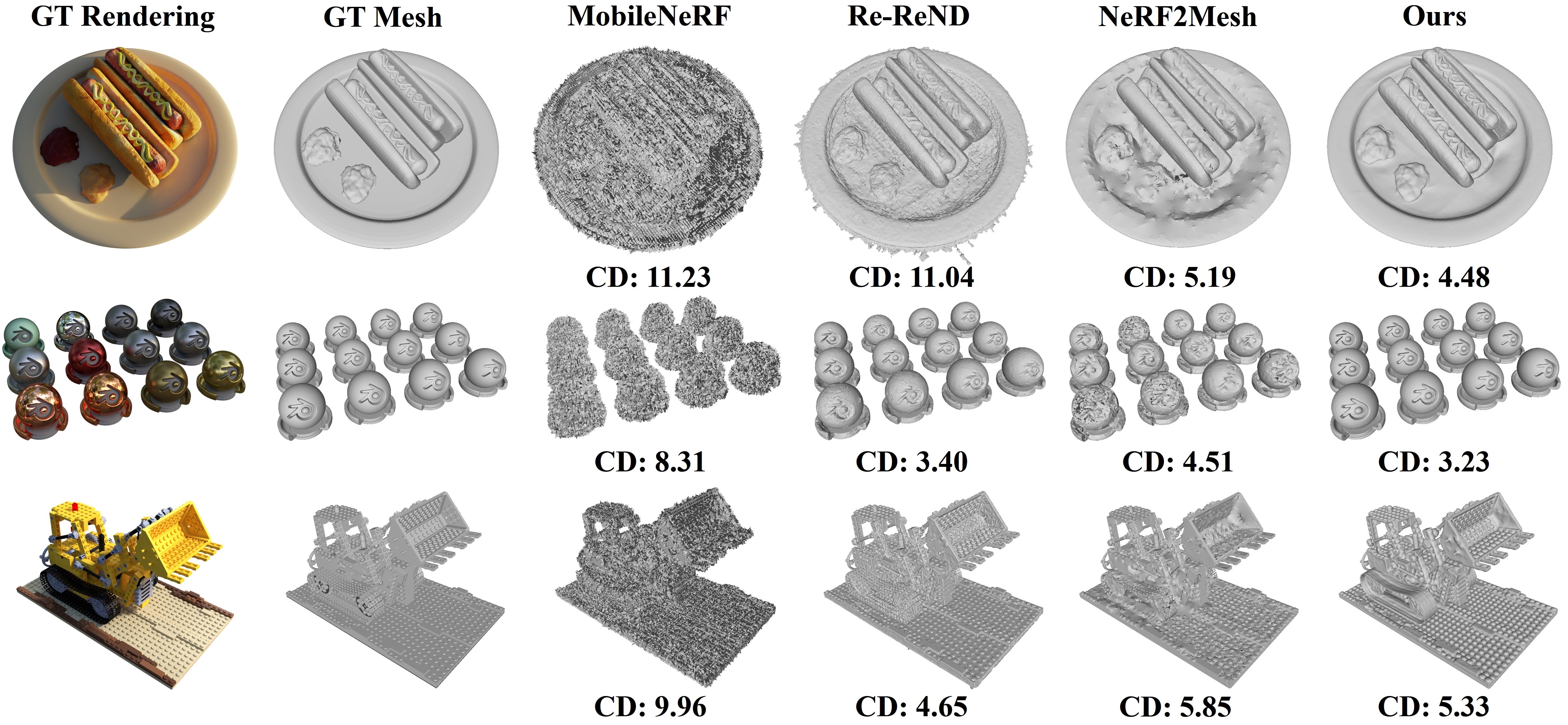}
    \caption{Comparison of meshes on NeRF synthetic dataset. Our method generates more accurate meshes than previous methods, especially for objects where view-dependent effects occur. The Chamfer Distance $\downarrow$ (the unit is $10^{-3}$) results are also provided.}
    \label{comp_mesh_new}
\end{figure*}

As depicted in Fig.~\ref{comp_mesh_new}, our model demonstrates the capability to reconstruct smoother surfaces for non-Lambertian objects (\textit{Materials}).
Additionally, we achieve more consistent mesh results when facing objects affected by environmental lighting conditions, such as \textit{Lego} and \textit{Hotdog}. The underlying reason for the improvements can be attributed to the explicit mesh with multi-view supervision employed for obtaining the geometry. Specifically, MobileNeRF~\cite{mobilenerf} synthesizes a mesh from the area constructed by interlacing polygons, leading to an inaccurate mesh with ambiguous boundaries. Re-ReND~\cite{rerend} and NeRF2Mesh~\cite{nerf2mesh} both train a NeRF-like model first and then extract mesh from the model. However, these NeRF-based models tend to interpret the view-dependent effects as virtual lights underneath the actual surfaces, leading to inaccurate mesh results. Meanwhile, by utilizing an explicit geometry, our method can learn a solid mesh with a smooth surface. Therefore, we surpass previous methods by generating accurate meshes for a greater variety of objects.\\
\noindent\textbf{Rendering Quality.} To evaluate the performance of rendering quality, we compare our results with the state-of-the-art methods that achieve real-time rendering on mobile devices~\cite{rerend,nerf2mesh,mobilenerf}. We also provide the results of the original NeRF model~\cite{MildenhallSTBRN20} and SNeRG~\cite{SNeRG} as reference. It is noteworthy that we set the texture resolution as 2048 in all models for fair comparison. We choose PSNR, SSIM, and LPIPS as the evaluation metrics. The experiments are conducted on the NeRF synthetic dataset~\cite{MildenhallSTBRN20}. As shown in Table~\ref{tab:comparison}, our method generates comparable results with previous methods. It is worth noting that our proposed method yields superior outcomes compared to previous approaches in terms of generating a smooth mesh. Moreover, our approach generates comparable results with methods that prioritize rendering over mesh quality.
\begin{table}[tb]
    \centering
    \caption{Quantitative comparison on NeRF synthetic dataset. We report PSNR, SSIM, and LPIPS of previous methods and ours on the NeRF Synthetic dataset. Note that ``better rendering" setting of NeRF2Mesh is denoted by ``R" and ``better mesh" by ``M".}
    \begin{tabular}{l|cccc}
    \hline & PSNR $\uparrow$ & SSIM $\uparrow$ & LPIPS $\downarrow$ \\
    \hline NeRF~\cite{MildenhallSTBRN20} & 31.01 & 0.947 & 0.081 \\
     \hline
     SNeRG~\cite{SNeRG} & 30.38 & 0.950 & 0.050 \\
     MobileNeRF~\cite{mobilenerf} & 30.90 & 0.947 & 0.062  \\
     Re-ReND~\cite{rerend} & 29.00 & 0.934 & 0.080 \\ 
     NeRF2Mesh(R)~\cite{nerf2mesh} & 31.04 & 0.948 & 0.066 \\
     NeRF2Mesh(M)~\cite{nerf2mesh} & 29.76 &  0.940 & 0.072 \\
    \hline Ours & 30.03 & 0.944& 0.070 \\
    \hline
    \end{tabular}
    \label{tab:comparison}
\end{table}

\noindent\textbf{Efficiency.} In terms of model efficiency, we compare the disk storage and the training time of our method and existing methods~\cite{mobilenerf,nerf2mesh,rerend}.
\begin{table}[tb]
    \centering
    \caption{Comparison of disk storage and training time. We report the size (MB) of the exported models, the training time (hour) and the rendering speed (FPS).}
    \resizebox{\linewidth}{!}{
    \begin{tabular}{l|ccc}
        \hline & Disk Storage $\downarrow$ & Training Time $\downarrow$ & Rendering Speed $\uparrow$\\
      \hline   MobileNeRF~\cite{mobilenerf} & 125.7 MB & 20-24 h & 42 FPS  \\
      \hline Re-ReND~\cite{rerend} & 198.4 MB & 55-60 h & 54 FPS \\
      \hline 2DGS~\cite{2dgs} & 78.8 MB & \textbf{0-1 h} & 50 FPS \\  
      \hline NeRF2Mesh~\cite{nerf2mesh} & 73.5 MB & 1-2 h & 52 FPS \\  
      \hline Ours & \textbf{46.9 MB} &  1-2 h & 50 FPS  \\
       \hline
    \end{tabular}}
    \label{tab:eng}
\end{table}
For a fair comparison, all the models are trained on a single Tesla-V100 (32G) and the disk storage consists of the uncompressed mesh file (.obj and .mtl), texture file (.png), and the final model file (.json). As shown in Table~\ref{tab:eng}, our method consumes smaller disk storage than previous methods. This can be attributed to the utilization of an accurate surface-based model, which prevents the generation of redundant vertices and faces. Additionally, our approach incorporates a lightweight learnable texture in the rendering process, further reducing the overall storage. As for the model training time, our method achieves a much faster training speed than MobileNeRF~\cite{mobilenerf} and Re-ReND~\cite{rerend}.
With the incorporation of the learnable implicit texture, our method directly optimizes the final texture required for rendering and bypasses the excessive computation required by these methods~\cite{mobilenerf, rerend} to transform the sampling features from the radiance field to the final textures.
For gaussian-based methods like 2DGS~\cite{2dgs}, although they can easily render at interactive framerates, they are not built for end-to-end mobile rendering. To further compare the rendering efficiency, we implement a \textit{three.js} based renderer for mobile rendering. It can be seen that 2DGS requires excessive disk storages for rendering while our method is more mobile-friendly considering the aspect of storages.
As for NeRF2Mesh~\cite{nerf2mesh}, since it utilizes a fast-training NeRF model before obtaining the mesh and texture, it trains slightly faster than our method. However, our model can handle view-dependent surfaces that NeRF2Mesh~\cite{nerf2mesh} could not with comparable training time. The rendering speed is measured using iPhone13 and the results are reported in Table~\ref{tab:eng}. It is noteworthy that the reported training time (1-2 h) includes both training time required for stage 1 and stage 2 and the time required for OpenMVS reconstruction.

Overall, EvaSurf exhibits high-fidelity rendering results and generates more accurate mesh than previous methods. Furthermore, from the perspective of real-world applications, EvaSurf is more suitable as it requires much less disk storage and training time, which could reduce the constraints for hardware, and can render realistic results with accurate meshes in real-time on mobile devices.

\begin{figure*}[tb]
    \centering
    \includegraphics[width=\linewidth]{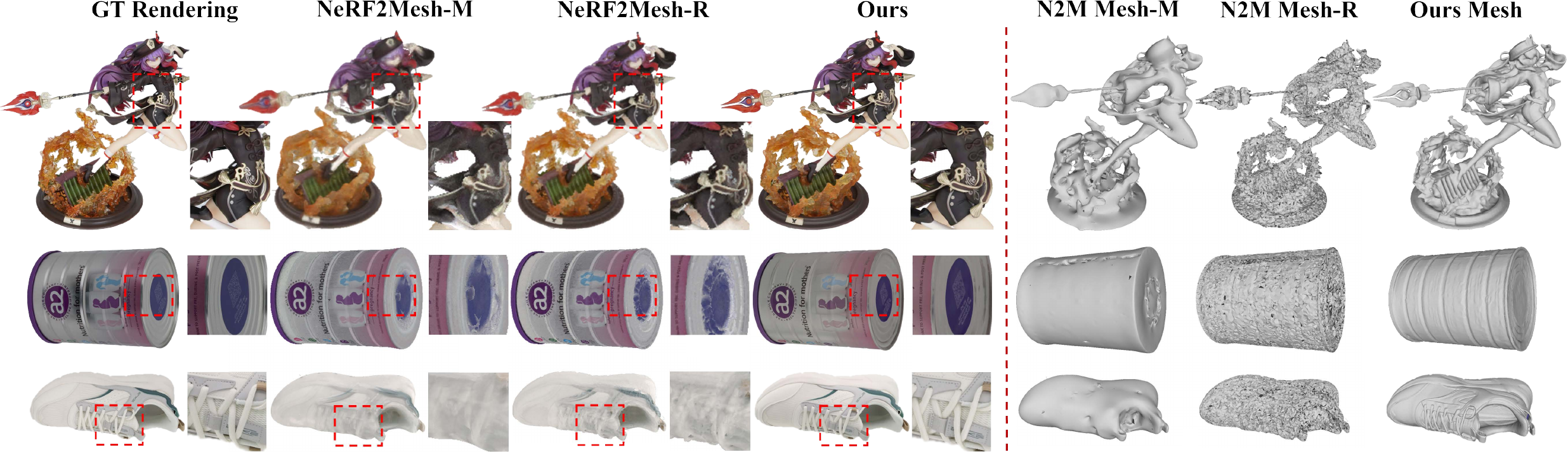}
    \caption{Results on real-world dataset. Our method can reconstruct 3D objects from the real-world dataset with high-fidelity rendered results and more accurate mesh than the existing methods. We denote the NeRF2Mesh~\cite{nerf2mesh} better mesh setting as ``N2M Mesh-M" and better rendering setting as ``N2M Mesh-R" in the figure. }
    \label{fig:real_world_examples}
\end{figure*}

\subsection{Comparisons on Real-world 3D Objects}
As depicted in Fig.~\ref{fig:real_world_examples}, we also compare our rendering results and mesh results with NeRF2Mesh~\cite{nerf2mesh} on a real-world dataset. We train NeRF2Mesh under two settings following the authors' instructions. We first train NeRF2Mesh under the ``better rendering" setting and extract the mesh named ``N2M Mesh-R" in the figure. Then we train it under the ``better mesh" settings and extract the mesh named as ``N2M Mesh-M", the renderings are generated correspondingly. 
From the perspective of \textbf{geometry quality}, the ``better mesh" setting of NeRF2Mesh generates a smoother surface than the ``better rendering" setting but fails to reconstruct complex structures. On the other hand, using the ``better rendering" setting and decimating the mesh size could obtain a better shape but the surface appears to be not smooth. This originates from the exportation methodology of NeRF-based model, which is incapable of reconstructing smooth complex surfaces. Contrarily, our method employs an explicit representation and utilizes a multi-view supervision module and a set of progressive grids to optimize the mesh. This allows our model to generate accurate surfaces effectively for these complex structures. 
From the perspective of \textbf{rendering quality}, our method achieves better appearances than NeRF2Mesh~\cite{nerf2mesh} under both the ``better rendering" setting and ``better mesh" setting. NeRF2Mesh directly bakes the appearance of the NeRF model into texture, which cannot render images with view-dependent effects. Meanwhile, we optimize a learnable implicit texture with view-aware encoding. By modeling view dependence with different fidelity in the feature space, our model can generate high-fidelity renderings. 

\subsection{Ablation Studies}
\begin{figure}[tb]
    \centering
    \includegraphics[width=\linewidth]{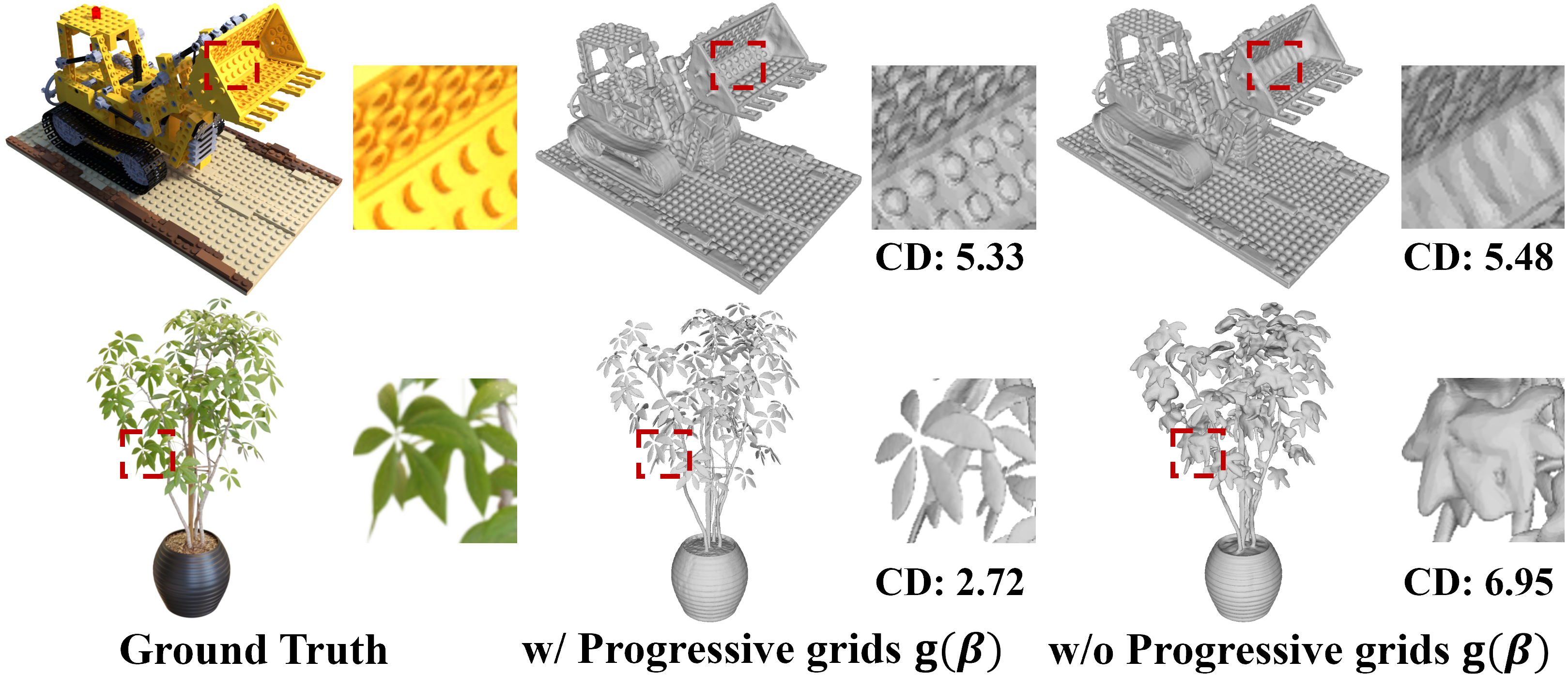}
    \caption{Ablation of the Progressive Grids. By utilizing a set of progressive grids, we could generate a mesh with better details. Chamfer Distance (CD $\downarrow$, unit is $10^{-3}$) is provided as reference.}
    \label{progressive}
\end{figure}

\noindent\textbf{Progressive Grids.}
Progressive grids with multi-resolution are employed in the surface-based model to realize the coarse-to-fine training strategy for better details.  The effectiveness of these progressive grids is demonstrated in Fig.~\ref{progressive}. We observe that a coarse shape is initialized when using low-resolution grids, while high-frequency details can be captured when using high-resolution grids. As demonstrated in Table~\ref{ablation_pg}, utilizing progressive grids in our surface-based model can lead to improved mesh quality.
\begin{table}[tb]
    \centering
     \caption{Comparison of Chamfer Distance (CD) results. We report the evaluation results of utilizing progressive grids or not in the surface-based model.}
    \begin{tabular}{c|c}
        \hline & CD $\downarrow$ (unit: $10^{-3}$)  \\
      \hline  w/ Progressive grids $g(\beta)$ & \textbf{4.39}   \\
      \hline w/o Progressive grids $g(\beta)$ & 5.35  \\
       \hline
    \end{tabular}
    \label{ablation_pg}
\end{table}

\begin{figure}[tb]
    \centering
    \includegraphics[width=0.85\linewidth]{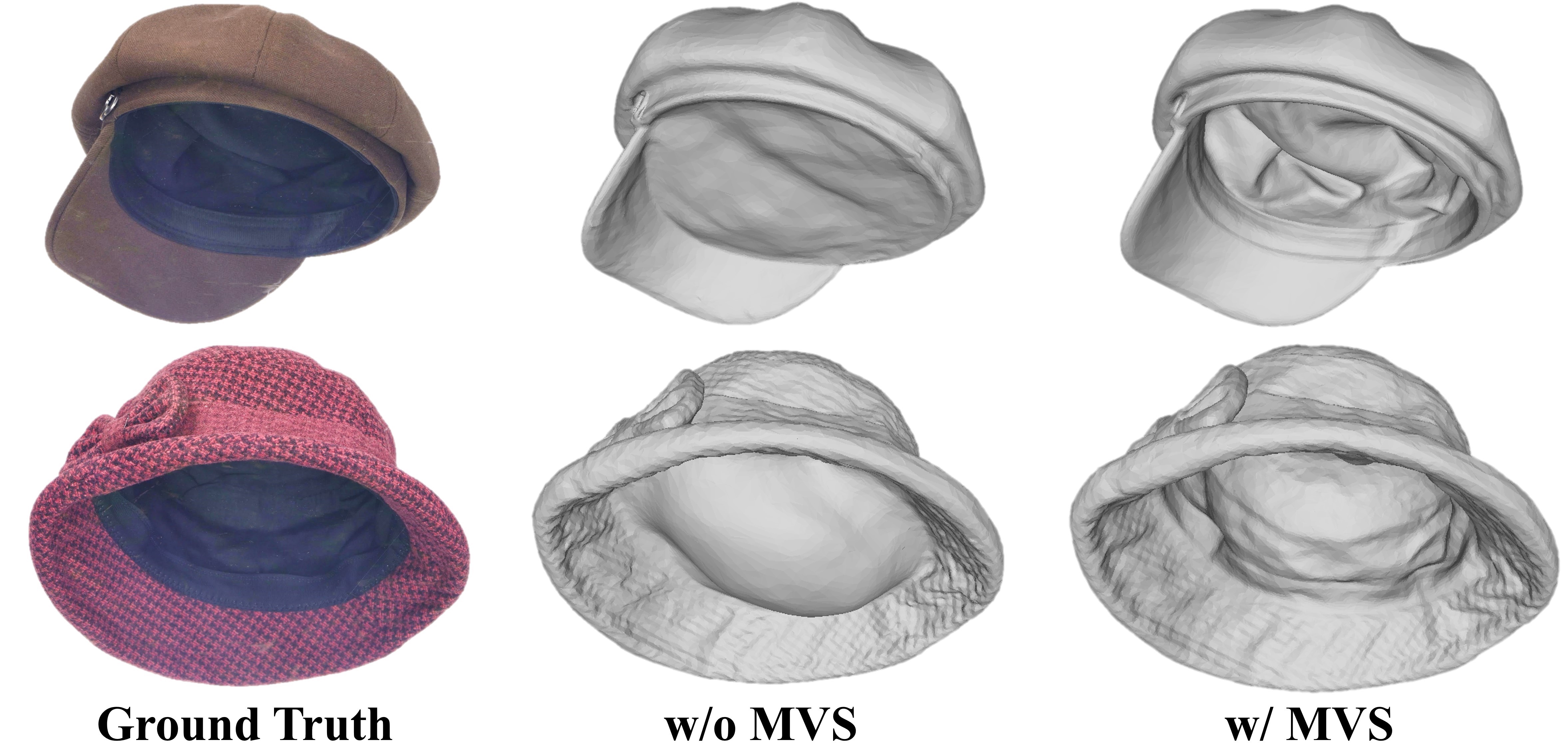}
    \caption{Ablation of the Multiview Supervision Module (MVS).
    Note that the Ground Truth figure is relighted for a better view of the object's inner structure. It could be seen that we could reconstruct the interior structure of the hat with the aid of MVS.}
    \label{comp_mesh}
\end{figure}

\noindent\textbf{Multi-view Supervision Module.}
By explicitly initializing the geometry with an additional multi-view supervision module, our method establishes a solid foundation for generating an accurate mesh. To better present the effectiveness of the multi-view supervision module, we compare the mesh results, shown in Fig.~\ref{comp_mesh}.
 For the same object, it could be seen that with the additional MVS, our method reconstructs a more accurate mesh with better details. The model without the additional MVS is not capable of reconstructing the mesh with complex structures such as the interior structure of a hat which contains several wrinkles. Meanwhile, with the additional supervision provided by MVS, our model manages to generate an accurate and well-structured mesh that matches the ground-truth images.

\begin{figure}[tb]
 \centering
 \includegraphics[width=0.8\linewidth]{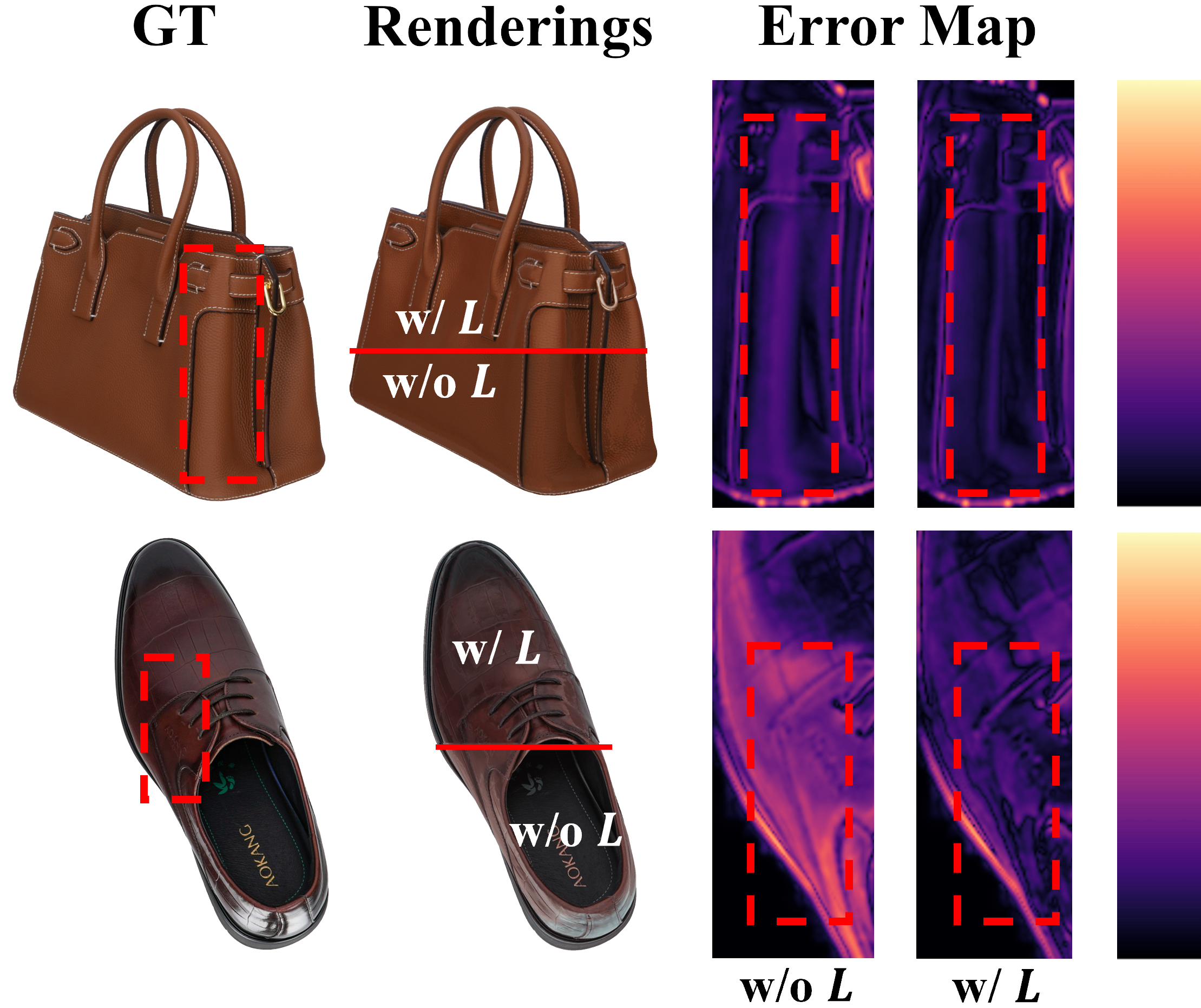}
 \caption{Ablation results of view-aware encoding on the real-world dataset. The renderings are stacked together for clearer comparison and the error maps are also provided. We use the color scale from FLIP~\cite{flip}.}
 \label{view_fig}
\end{figure}

\noindent\textbf{View-aware Encoding.}
In our 3D object reconstruction method, we optimize a 12-dimensional learnable implicit texture to render high-fidelity images on various devices. To incorporate the view-dependent condition during optimization, we apply view-aware encoding to the original implicit texture. By modeling the view dependence with different fidelity, several view-dependent effects can be rendered by our method. Comparison results on the real-world dataset are displayed in Fig.~\ref{view_fig} and the evaluation metrics are also reported in Table~\ref{ablation}. It could be seen that we obtain better renderings with the set of Gaussians. Since some regions are observable from only a limited range of viewing directions, modeling the view dependence with the same fidelity would lead to an average appearance aggregated from input views. By utilizing the Gaussians in view-aware encoding, we model the view-dependent information with different fidelity. By assigning different weights to each dimension of the texture in the feature space, we enable the model to reflect more realistic results. Furthermore, we fix the geometry reconstructed from stage 1 and employ the spherical gaussians used in BakedSDF~\cite{bakedsdf} and the MLPs in NeRF2Mesh~\cite{nerf2mesh} to further demonstrate the effectiveness of our anisotropy modeling strategies. The results shown in Table~\ref{ablation} proves that our feature-level Gaussians outperforms other anisotropy modelling strategies. 

\begin{table}[tb]
    \centering
    \caption{Ablation of view-aware encoding. The view-aware encoded model exhibits better view-dependent rendering results and is more effective than other anisotropy modeling strategy}.
    \begin{tabular}{c|ccc}
        \hline & PSNR $\uparrow$ & SSIM $\uparrow$ & LPIPS $\downarrow$ \\
      \hline  w/ $L(f,v_{cos})$ & \textbf{27.58} & \textbf{0.891}  & \textbf{0.208} \\
      \hline w/ SG Modeling~\cite{bakedsdf} & 27.24 & 0.889 & 0.209 \\ 
      \hline w/ MLP Modeling~\cite{nerf2mesh} & 26.96 & 0.887 & 0.212 \\
        \hline w/o $L(f,v_{cos})$ & 26.69 & 0.884 & 0.215 \\ 
       \hline
    \end{tabular}
    \label{ablation}
\end{table}

Our method uses feature-level Gaussians to model anisotropy and set the hyperparametr $\alpha$ in Eq.~\ref{feature} as 0.4 in experiments. This hyperparameter models the variance of the feature-level Gaussians, and the value should be determined carefully to ensure the model can capture appropriate range of anisotropy. Therefore, we present the ablation experiments on $\alpha$ and the rendering quality in Table~\ref{alpha}. It can be seen that either too large or too small would lead to inferior rendering results, and we determine the value through experiments and find out that the optimal value lies in around [0.4, 0.45], slightly leaning towards 0.4. Therefore, we chose $\alpha$ to be 0.4 by default.
\begin{table}[tb]
    \centering
    \caption{Ablation on the hyperparameter $\alpha$. We present the evaluation of PSNR by adjusting the value of $\alpha$}
    \begin{tabular}{c|ccccccc}
        \hline $\alpha$ & 0.1 & 0.2 & 0.4 & 0.6 & 0.8 & 1.0 & 1.4 \\
        \hline  PSNR ($\uparrow$) & 26.64 & 26.85 &27.58 & 27.39 & 26.42 & 26.04 & 25.96\\
        \hline
    \end{tabular}
    \label{alpha}
\end{table}

We also provide the detailed evaluation results for the ablation of the view-aware encoding utilized to model view-dependent effects. The PSNR, SSIM, and LPIPS results for each case of the real-world dataset are reported in Table~\ref{gaussian}. We also reported the differences $\Delta$PSNR, $\Delta$SSIM and $\Delta$LPIPS in the table to demonstrate the effectivenss. It is noteworthy that by utilizing the view-aware encoding, we obtained better results in all the metrics except for a slight decrease of SSIM and LPIPS for Object\_11.
\begin{table*}[tb]
    \centering
    \caption{Per-scene Evaluation for the real-world dataset. We report PSNR $\uparrow$, SSIM $\uparrow$, and LPIPS $\downarrow$ for each case in the real-world dataset. The differences $\Delta$PSNR, $\Delta$SSIM, and $\Delta$LPIPS are also reported. }
    \begin{tabular}{c|c|cc|cc|cc}
         \hline \textbf{Object} & \begin{tabular}[c]{@{}l@{}}View-aware\\ Encoding $L$ \end{tabular}& PSNR $\uparrow$ & $\Delta$PSNR &  SSIM $\uparrow$ & $\Delta$SSIM  & LPIPS $\downarrow$ & $\Delta$LPIPS \\
        \hline  01 & w & 26.56 & +0.17 & 0.899 & +0.003 &0.162 & -0.005 \\
                   & w/o &26.39 & - & 0.896 & - &0.167 & -  \\
                02 & w & 24.70 & +0.22 & 0.860 &+0.003 & 0.180  & -0.001 \\
                 & w/o & 24.48 & - & 0.857 & - & 0.181 & - \\
               03 & w & 29.25 & +0.50 & 0.952 & +0.002 & 0.161 & -0.002  \\
               & w/o & 28.75 & - & 0.950 & - & 0.163 & - \\
                04 & w & 25.84 & +0.88 & 0.869 &+0.015 & 0.219 & -0.008 \\
                 & w/o & 24.96 & - & 0.854 & - &  0.227 & - \\
               05 & w & 29.33 & +0.83 & 0.931 & +0.006 & 0.125 & -0.013 \\
               & w/o & 28.50 & - & 0.925 & - & 0.138 & -   \\
              06 & w & 29.59 & +0.22 & 0.942 & +0.001 & 0.187 & -0.001 \\
                & w/o & 29.37 & - & 0.941 & - &  0.186 & -\\
               07 & w & 26.68 & +1.36 & 0.898 & +0.114 & 0.217 &  -0.010 \\
               & w/o & 25.32 & - & 0.884 & - & 0.227 & -   \\
               08 & w & 23.90 & +0.81 & 0.880 & +0.004 & 0.249 & -0.002 \\
               & w/o & 23.09 & - & 0.876 & - & 0.251 & - \\
               09 & w & 29.55 & +1.62 & 0.911 & +0.010 & 0.201 & -0.012  \\
               & w/o & 27.93 & - & 0.901 & - & 0.213 & -    \\
               10 & w & 29.69 & +1.28 & 0.859 & +0.005& 0.300 & -0.002 \\
               & w/o & 28.41 & - & 0.854& -  & 0.302& -  \\
               11 & w & 28.11 & +0.81 & 0.792 & -0.008 & 0.310 & +0.011  \\
               & w/o & 27.30 & - & 0.800& -  & 0.299  & -   \\
               12 & w & 26.71 & +2.12 & 0.884 & +0.021 & 0.197 &-0.015 \\
               & w/o & 24.59 & - & 0.863 & - & 0.212 & - \\
               13 & w & 26.21 & +1.00  & 0.892 & +0.011 & 0.198 & -0.014  \\
               & w/o & 25.21 & - & 0.881 & - & 0.212 & -   \\
               14 & w & 27.38 & +0.98 & 0.853 & + 0.018 & 0.197 & -0.014 \\
               & w/o & 26.40 & - & 0.835 & - & 0.211 & - \\
               15 & w & 30.13 & +0.52 & 0.942 & +0.007 & 0.218 & -0.013   \\
               & w/o & 29.61 & - & 0.935 & - & 0.231& -  \\
        \hline    
    \end{tabular}
    \label{gaussian}
\end{table*}

\noindent\textbf{Implicit Texture.} In our model, we store the 12-dimensional implicit texture as 3 \textit{PNG} in the rendering package. We then conduct experiments on using 8-dimensional implicit texture and 16-dimensional implicit texture in our model, which are stored as 2 \textit{PNG} and 4 \textit{PNG} correspondingly. The evaluation results of PSNR, SSIM, LPIPS, and the final disk storages are reported in Table~\ref{channel}. It could be seen that we obtain improved renderings by increasing the channel number but consume more disk storage. To strike a balance between the rendering quality and disk storage, we choose 12 as the default channel number of the implicit texture. The channel number could be decreased if users have more restrictions on the disk storage and increased if users require better renderings.
\begin{table}[tb]
    \centering
    \caption{Ablation of texture channel number. We report PSNR, SSIM, LPIPS and the disk storage for reference}
    \begin{tabular}{c|cccc}
        \hline Channel & PSNR $\uparrow$ & SSIM $\uparrow$ & LPIPS $\downarrow$ & Disk Storage $\downarrow$ \\
      \hline  12 & 27.58 & 0.891  & 0.208 & 16.5 MB \\
       8 & 27.28 & 0.887 & 0.213 & 11.2 MB \\
       16 & 27.78 & 0.893 & 0.205 & 22.1 MB \\
       \hline
    \end{tabular}
    \label{channel}
\end{table}

In addition, we use 2048 as the texture resolution by default, and we conducted ablation experiment on the texture resolution using 1024, 2048 and 4096 as the resolution. The evaluation results of PSNR, disk storages and rendering speed are presented in Table~\ref{res}. It can be seen that there exists trade-offs between the rendering performance and the texture resolution. By choosing 2048 as the resolution, we obtain less disk storages without compromising rendering quality and rendering speed too much.
\begin{table}[tb]
    \centering
    \caption{Ablation of texture resolution. We report PSNR, disk storages and rendering speed for reference}
    \begin{tabular}{c|ccc}
        \hline Resolution & PSNR $\uparrow$ & Disk Storage $\downarrow$ & Rendering Speed $\uparrow$ \\
       \hline  2048 & 27.58 & 16.5 MB & 50 FPS\\
       1024 & 27.06 & 14.4 MB & 54 FPS\\
       4096 & 27.74 & 19.8 MB & 44 FPS\\
       \hline
    \end{tabular}
    \label{res}
\end{table}

\subsection{Additional Results}
\begin{figure}[tb]
    \centering
    \includegraphics[width=\linewidth]{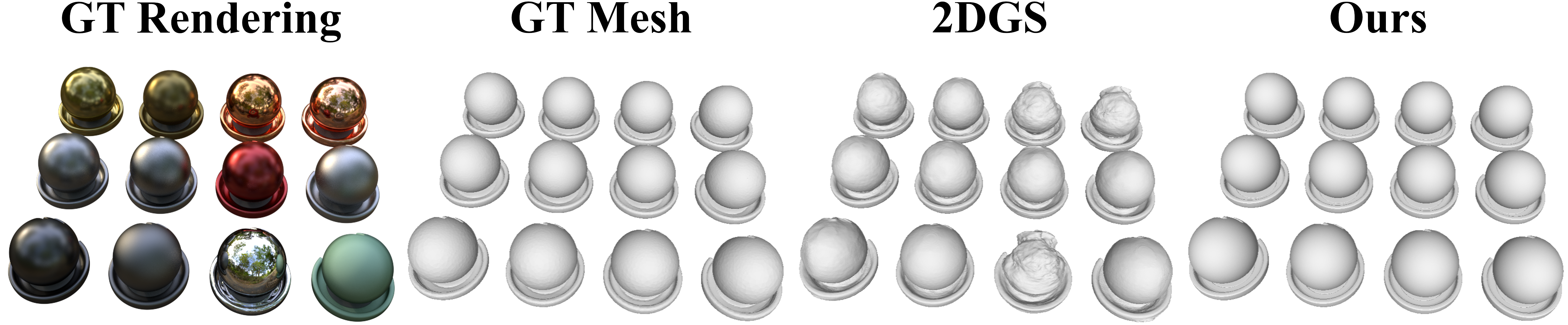}
    \caption{Mesh comparison results with state-of-the-art Gaussian-based methods. Our methods demonstrates more accurate results on reflective objects.}
    \label{2dg}
\end{figure}

\noindent\textbf{Comparison with Gaussian-based methods.}
Gaussian-based methods~\cite{3dgs,Yu2023MipSplatting,scaffoldgs, r3dgs} cannot render on mobile devices due to requirement of large disk storage and lack of mobile-end optimization. Besides, the reconstructed geometry is not satisfied for further application. 2DGS~\cite{2dgs} can reconstruct the corresponding geometry but is still unable to render on mobile devices directly. In contrast, EvaSurf can achieve real-time rendering on mobile devices while reconstructing accurate geometry for further applications. We compare the reconstructed geometry of our method and 2DGS~\cite{2dgs} in Figure~\ref{2dg} and Table~\ref{mesh_results}. It can be seen that 2DGS cannot handle objects with reflective surfaces while EvaSurf reconstruct highly accurate geometry and achieve better results for all the synthetic scenes.

\begin{figure}[tb]
    \centering
    \includegraphics[width=\linewidth]{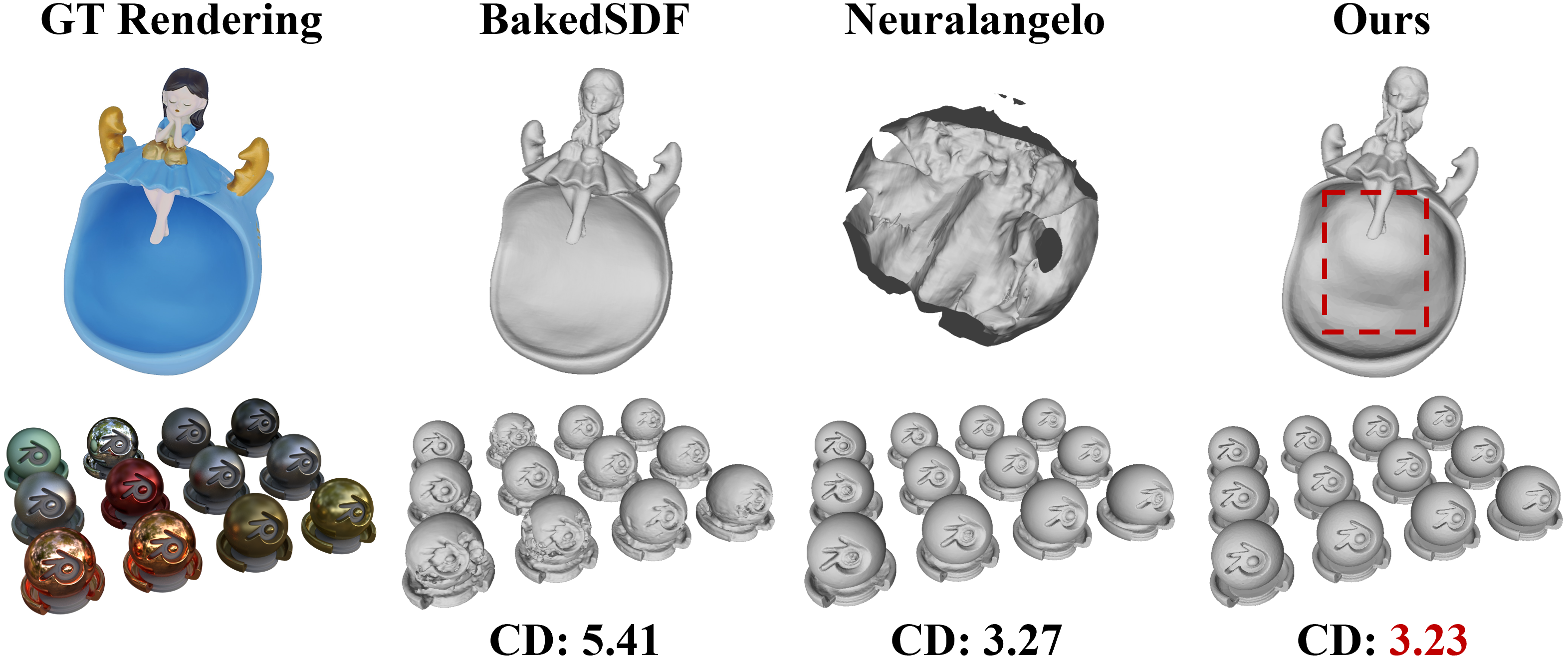}
    \caption{Mesh comparison results with state-of-the-art SDF-based methods. The Chamfer Distance (CD $\downarrow$, unit is $10^{-3}$) for the \textit{materials} scene is provided as reference.}
    \label{mesh_comp_re}
\end{figure}

\begin{figure}[tb]
    \centering
    \includegraphics[width=\linewidth]{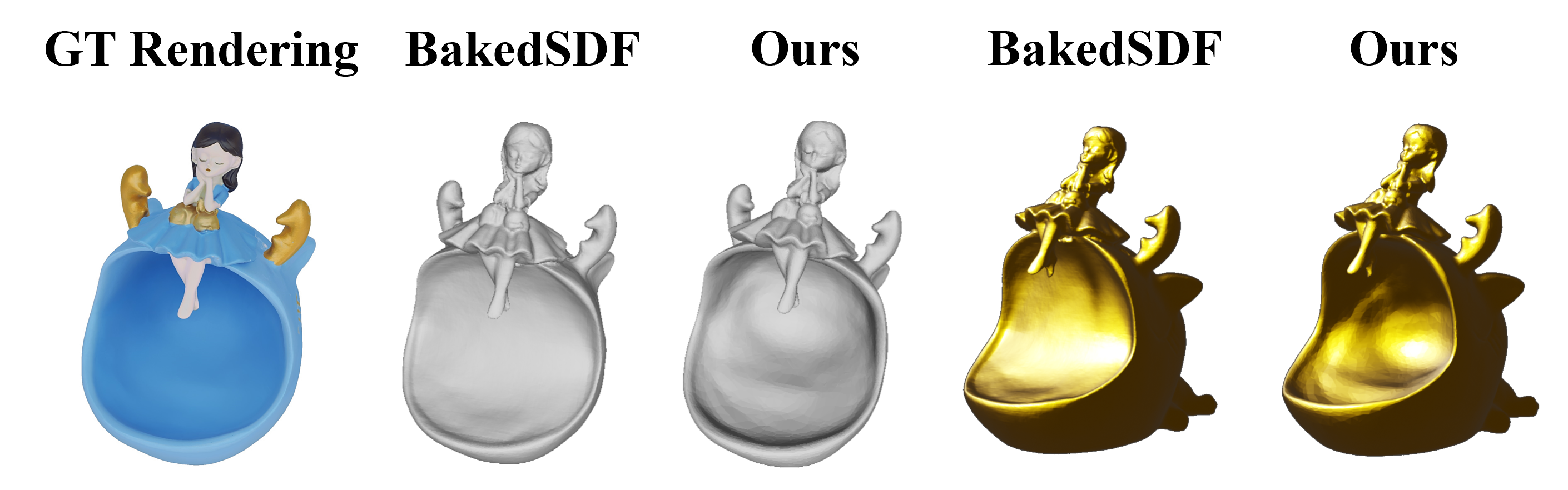}
    \caption{Mesh comparison results in application. We use Blender to edit the material and perform relighting.}
    \label{blender}
\end{figure}

\noindent\textbf{Comparison with SDF-based methods.}
As shown in Fig.~\ref{mesh_comp_re}, BakedSDF~\cite{bakedsdf} can not reconstruct accurate shapes for complex objects and reflective objects. Neuralangelo~\cite{DBLP:conf/cvpr/Li0ETU0L23} requires excessive training time($\sim$27h) and does not support real-time rendering. EvaSurf reconstructs \textbf{more accurate geometry} with much smaller sizes and allows real-time rendering on mobile devices. 
From the perspective of application, the accurate shape of the mesh is crucial and affects the results of downstream tasks like material editing and relighting. As shown in Fig.~\ref{blender}, we use Blender to edit the material and perform relighting. It can be seen that with the accurate shape, the mesh generated by our method could be directly applied to material editing or relighting using modern software without sacrificing the actual structure of the object.

\noindent\textbf{Rendering Memory Footprint Comparison.}
We further provide the comparison of rendering memory footprint of different methods to demonstrate the applicability of EvaSurf. We use the NeRF synthetic dataset as the example and the results are shown in Table~\ref{mf}. It is noteworthy that we implement a \textit{three.js} based renderer for 2DGS~\cite{2dgs} for mobile rendering. It can be seen that our method requires much less memory footprint when rendering at interactive frame rates. This can be beneficial to extending the applications of 3D reconstruction models in real-life scenarios.
\begin{table}[tb]
    \centering
    \caption{Comparison of rendering memory footprint. We present the rendering memory footprint of different methods.}
    \begin{tabular}{c|cc}
        \hline Method  & Memory Footprint ($\downarrow$) & Rendering Speed ($\uparrow$) \\
        \hline Re-ReND~\cite{rerend} & 128 MB & 54 FPS\\
        \hline MobileNeRF~\cite{mobilenerf} & 199 MB & 42 FPS\\
        \hline 2DGS~\cite{2dgs} &  82 MB & 50 FPS\\
        \hline NeRF2Mesh~\cite{nerf2mesh} & 75 MB & 52 FPS\\
        \hline Ours & \textbf{48 MB} &50 FPS\\
        \hline
    \end{tabular}
    \label{mf}
\end{table}

\begin{figure}[tb]
    \centering
    \includegraphics[width=0.9\linewidth]{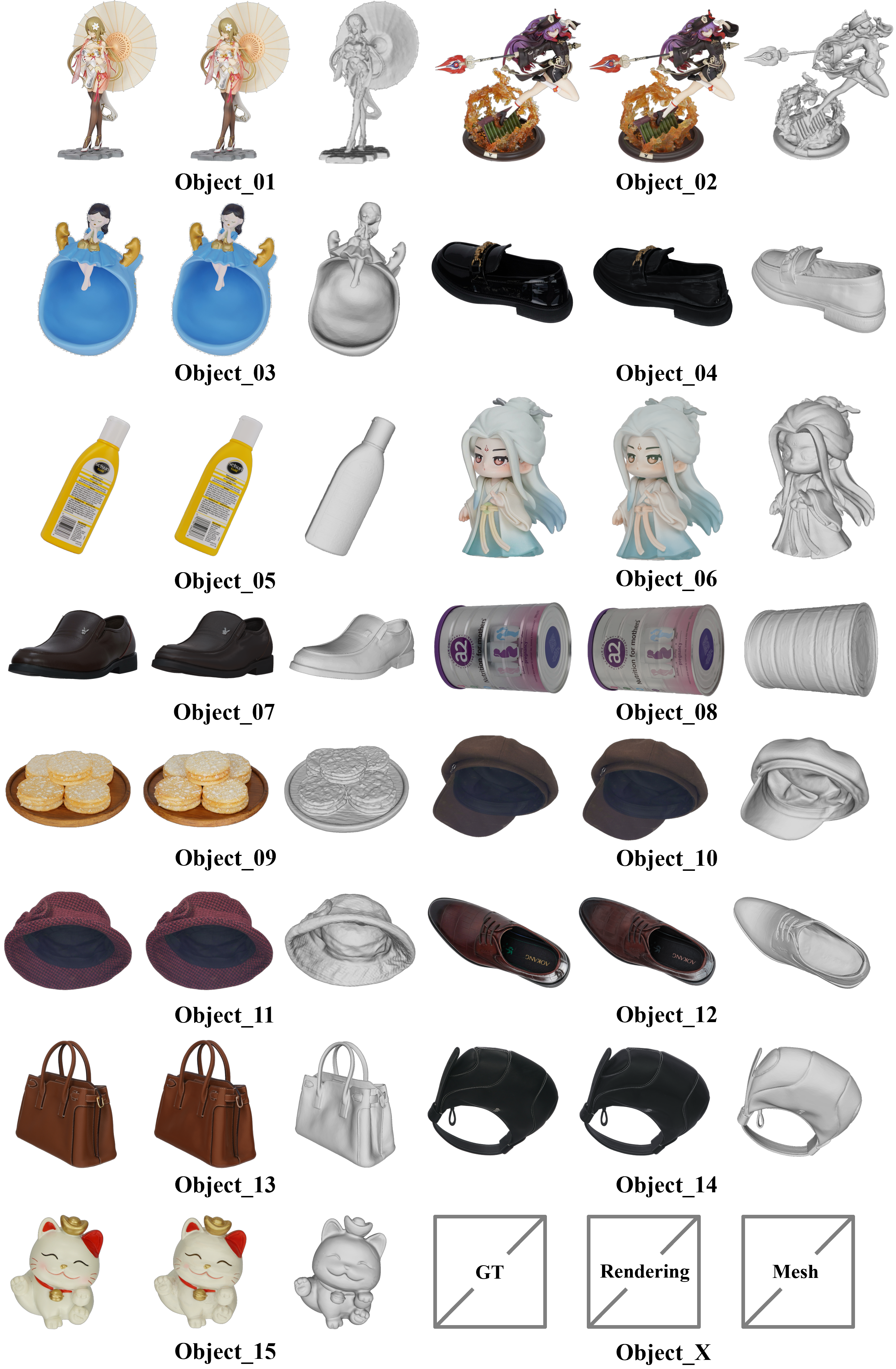}
    \caption{Reconstruction results of the real-world dataset. We showcase all the reconstruction results of the real-world dataset we used in the experiments}
    \label{recon}
\end{figure}

\noindent\textbf{Real-world Dataset Reconstruction Results.}
We provide the reconstruction results of the real-world dataset including the rendering results and mesh results. The reconstruction results are shown in Fig.~\ref{recon}. 

\section{Conclusion}
In conclusion, we propose a novel method, EvaSurf, an efficient view-aware implicit textured surface reconstruction method on mobile devices. We first introduce a multi-view supervision to the surface-based model for the reconstruction of accurate mesh.
We then equip a learnable implicit texture with view-aware encoding to render high-fidelity images with view-dependent effects.
A neural shader, which is a lightweight MLP, is further employed to enable real-time rendering on mobile devices. 
Our approach is not limited to synthetic data but also manages to render real-world objects with high-quality appearance and accurate mesh. 
With the power of fast training and small disk storage, our method is capable of rendering on mobile devices at interactive frame rates without compromising on the quality of the output, generalized for daily applications.

\section{Limitation}
Our limitation lies in thin or furry structures which is a common challenge for SDF-based methods. Although we use progressive spatial grids to reconstruct complex geometry, thin structures like branches of a Ficus still lack high-frequency details. This limitation can be released by bringing in volume components.

\section{Acknowledgements}
This work was supported in part by NSFC (62201342), and Shanghai Municipal Science and Technology Major Project (2021SHZDZX0102). Authors would like to appreciate the Student Innovation Center of SJTU for providing GPUs.

\newpage

\bibliographystyle{IEEEtran}
\bibliography{ref}

% Generated by IEEEtran.bst, version: 1.14 (2015/08/26)
\begin{thebibliography}{10}
\providecommand{\url}[1]{#1}
\csname url@samestyle\endcsname
\providecommand{\newblock}{\relax}
\providecommand{\bibinfo}[2]{#2}
\providecommand{\BIBentrySTDinterwordspacing}{\spaceskip=0pt\relax}
\providecommand{\BIBentryALTinterwordstretchfactor}{4}
\providecommand{\BIBentryALTinterwordspacing}{\spaceskip=\fontdimen2\font plus
\BIBentryALTinterwordstretchfactor\fontdimen3\font minus \fontdimen4\font\relax}
\providecommand{\BIBforeignlanguage}[2]{{%
\expandafter\ifx\csname l@#1\endcsname\relax
\typeout{** WARNING: IEEEtran.bst: No hyphenation pattern has been}%
\typeout{** loaded for the language `#1'. Using the pattern for}%
\typeout{** the default language instead.}%
\else
\language=\csname l@#1\endcsname
\fi
#2}}
\providecommand{\BIBdecl}{\relax}
\BIBdecl

\bibitem{AndersenVP19}
D.~Andersen, P.~Villano, and V.~Popescu, ``{AR} {HMD} guidance for controlled hand-held 3d acquisition,'' \emph{{IEEE} Trans. Vis. Comput. Graph.}, vol.~25, no.~11, pp. 3073--3082, 2019.

\bibitem{HuangDGN17}
J.~Huang, A.~Dai, L.~J. Guibas, and M.~Nie{\ss}ner, ``3dlite: towards commodity 3d scanning for content creation,'' \emph{{ACM} Trans. Graph.}, vol.~36, no.~6, pp. 203:1--203:14, 2017.

\bibitem{LiuCKKH21}
Z.~Liu, Y.~Cao, Z.~Kuang, L.~Kobbelt, and S.~Hu, ``High-quality textured 3d shape reconstruction with cascaded fully convolutional networks,'' \emph{{IEEE} Trans. Vis. Comput. Graph.}, vol.~27, no.~1, pp. 83--97, 2021.

\bibitem{LombardiSSSLS19}
S.~Lombardi, T.~Simon, J.~M. Saragih, G.~Schwartz, A.~M. Lehrmann, and Y.~Sheikh, ``Neural volumes: learning dynamic renderable volumes from images,'' \emph{{ACM} Trans. Graph.}, vol.~38, no.~4, pp. 65:1--65:14, 2019.

\bibitem{LoubetHJ19}
G.~Loubet, N.~Holzschuch, and W.~Jakob, ``Reparameterizing discontinuous integrands for differentiable rendering,'' \emph{{ACM} Trans. Graph.}, vol.~38, no.~6, pp. 228:1--228:14, 2019.

\bibitem{LuanZBD21}
F.~Luan, S.~Zhao, K.~Bala, and Z.~Dong, ``Unified shape and {SVBRDF} recovery using differentiable monte carlo rendering,'' \emph{Comput. Graph. Forum}, vol.~40, no.~4, pp. 101--113, 2021.

\bibitem{LyuWLC020}
J.~Lyu, B.~Wu, D.~Lischinski, D.~Cohen{-}Or, and H.~Huang, ``Differentiable refraction-tracing for mesh reconstruction of transparent objects,'' \emph{{ACM} Trans. Graph.}, vol.~39, no.~6, pp. 195:1--195:13, 2020.

\bibitem{Martin-BruallaR21}
R.~Martin{-}Brualla, N.~Radwan, M.~S.~M. Sajjadi, J.~T. Barron, A.~Dosovitskiy, and D.~Duckworth, ``Nerf in the wild: Neural radiance fields for unconstrained photo collections,'' in \emph{{CVPR}}, 2021, pp. 7210--7219.

\bibitem{MildenhallSTBRN20}
B.~Mildenhall, P.~P. Srinivasan, M.~Tancik, J.~T. Barron, R.~Ramamoorthi, and R.~Ng, ``Nerf: Representing scenes as neural radiance fields for view synthesis,'' in \emph{{ECCV}}, 2020.

\bibitem{Niemeyer021}
M.~Niemeyer and A.~Geiger, ``{GIRAFFE:} representing scenes as compositional generative neural feature fields,'' in \emph{{CVPR}}, 2021, pp. 11\,453--11\,464.

\bibitem{NiemeyerMOG20}
M.~Niemeyer, L.~M. Mescheder, M.~Oechsle, and A.~Geiger, ``Differentiable volumetric rendering: Learning implicit 3d representations without 3d supervision,'' in \emph{{CVPR}}, 2020, pp. 3501--3512.

\bibitem{PumarolaCPM21}
A.~Pumarola, E.~Corona, G.~Pons{-}Moll, and F.~Moreno{-}Noguer, ``D-nerf: Neural radiance fields for dynamic scenes,'' in \emph{{CVPR}}, 2021, pp. 10\,318--10\,327.

\bibitem{wang2021nerfmm}
Z.~Wang, S.~Wu, W.~Xie, M.~Chen, and V.~A. Prisacariu, ``Nerf--: Neural radiance fields without known camera parameters,'' \emph{arXiv preprint arXiv:2102.07064}, 2021.

\bibitem{YuYTK21}
A.~Yu, V.~Ye, M.~Tancik, and A.~Kanazawa, ``pixelnerf: Neural radiance fields from one or few images,'' in \emph{{CVPR}}, 2021, pp. 4578--4587.

\bibitem{DBLP:conf/cvpr/VerbinHMZBS22}
D.~Verbin, P.~Hedman, B.~Mildenhall, T.~E. Zickler, J.~T. Barron, and P.~P. Srinivasan, ``Ref-nerf: Structured view-dependent appearance for neural radiance fields,'' in \emph{CVPR}, 2022, pp. 5481--5490.

\bibitem{nerv2021}
P.~P. Srinivasan, B.~Deng, X.~Zhang, M.~Tancik, B.~Mildenhall, and J.~T. Barron, ``Nerv: Neural reflectance and visibility fields for relighting and view synthesis,'' in \emph{CVPR}, 2021, pp. 7495--7504.

\bibitem{DBLP:journals/tog/ZhangSDDFB21}
X.~Zhang, P.~P. Srinivasan, B.~Deng, P.~E. Debevec, W.~T. Freeman, and J.~T. Barron, ``Nerfactor: neural factorization of shape and reflectance under an unknown illumination,'' \emph{{ACM} Trans. Graph.}, vol.~40, no.~6, pp. 237:1--237:18, 2021.

\bibitem{Barron_2023_ICCV}
J.~T. Barron, B.~Mildenhall, D.~Verbin, P.~P. Srinivasan, and P.~Hedman, ``Zip-nerf: Anti-aliased grid-based neural radiance fields,'' in \emph{ICCV}, 2023, pp. 19\,697--19\,705.

\bibitem{9909994}
G.-W. Yang, W.-Y. Zhou, H.-Y. Peng, D.~Liang, T.-J. Mu, and S.-M. Hu, ``Recursive-nerf: An efficient and dynamically growing nerf,'' \emph{IEEE Transactions on Visualization and Computer Graphics}, vol.~29, no.~12, pp. 5124--5136, 2023.

\bibitem{peng2023intrinsicngp}
B.~Peng, J.~Hu, J.~Zhou, X.~Gao, and J.~Zhang, ``Intrinsicngp: Intrinsic coordinate based hash encoding for human nerf,'' \emph{IEEE Transactions on Visualization and Computer Graphics}, 2023.

\bibitem{10144678}
C.~Wang, R.~Jiang, M.~Chai, M.~He, D.~Chen, and J.~Liao, ``Nerf-art: Text-driven neural radiance fields stylization,'' \emph{IEEE Transactions on Visualization and Computer Graphics}, no.~01, pp. 1--15, 5555.

\bibitem{DBLP:conf/iccv/ParkSBBGSM21}
K.~Park, U.~Sinha, J.~T. Barron, S.~Bouaziz, D.~B. Goldman, S.~M. Seitz, and R.~Martin{-}Brualla, ``Nerfies: Deformable neural radiance fields,'' in \emph{ICCV}, 2021, pp. 5845--5854.

\bibitem{DBLP:journals/corr/abs-2010-07492}
K.~Zhang, G.~Riegler, N.~Snavely, and V.~Koltun, ``Nerf++: Analyzing and improving neural radiance fields,'' \emph{arXiv preprint arXiv:2010.07492}, 2020.

\bibitem{SNeRG}
P.~Hedman, P.~P. Srinivasan, B.~Mildenhall, J.~T. Barron, and P.~E. Debevec, ``Baking neural radiance fields for real-time view synthesis,'' in \emph{{ICCV}}, 2021, pp. 5855--5864.

\bibitem{mobilenerf}
Z.~Chen, T.~A. Funkhouser, P.~Hedman, and A.~Tagliasacchi, ``Mobilenerf: Exploiting the polygon rasterization pipeline for efficient neural field rendering on mobile architectures,'' in \emph{CVPR}, 2023, pp. 16\,569--16\,578.

\bibitem{2dgs}
B.~Huang, Z.~Yu, A.~Chen, A.~Geiger, and S.~Gao, ``2d gaussian splatting for geometrically accurate radiance fields,'' in \emph{SIGGRAPH}, 2024.

\bibitem{3dgs}
B.~Kerbl, G.~Kopanas, T.~Leimk{\"{u}}hler, and G.~Drettakis, ``3d gaussian splatting for real-time radiance field rendering,'' \emph{{ACM} Trans. Graph.}, vol.~42, no.~4, pp. 139:1--139:14, 2023.

\bibitem{scaffoldgs}
T.~Lu, M.~Yu, L.~Xu, Y.~Xiangli, L.~Wang, D.~Lin, and B.~Dai, ``Scaffold-gs: Structured 3d gaussians for view-adaptive rendering,'' \emph{CVPR}, 2024.

\bibitem{Yu2023MipSplatting}
Z.~Yu, A.~Chen, B.~Huang, T.~Sattler, and A.~Geiger, ``Mip-splatting: Alias-free 3d gaussian splatting,'' \emph{CVPR}, 2024.

\bibitem{rerend}
S.~Rojas, J.~Zarzar, J.~C. P\'erez, A.~Sanakoyeu, A.~Thabet, A.~Pumarola, and B.~Ghanem, ``Re-rend: Real-time rendering of nerfs across devices,'' in \emph{ICCV}, 2023, pp. 3632--3641.

\bibitem{nerf2mesh}
J.~Tang, H.~Zhou, X.~Chen, T.~Hu, E.~Ding, J.~Wang, and G.~Zeng, ``Delicate textured mesh recovery from nerf via adaptive surface refinement,'' in \emph{ICCV}, 2023, pp. 17\,739--17\,749.

\bibitem{bakedsdf}
L.~Yariv, P.~Hedman, C.~Reiser, D.~Verbin, P.~P. Srinivasan, R.~Szeliski, J.~T. Barron, and B.~Mildenhall, ``Bakedsdf: Meshing neural sdfs for real-time view synthesis,'' in \emph{SIGGRAPH}, E.~Brunvand, A.~Sheffer, and M.~Wimmer, Eds., 2023, pp. 46:1--46:9.

\bibitem{DBLP:journals/tog/ReiserSVSMGBH23}
C.~Reiser, R.~Szeliski, D.~Verbin, P.~P. Srinivasan, B.~Mildenhall, A.~Geiger, J.~T. Barron, and P.~Hedman, ``{MERF:} memory-efficient radiance fields for real-time view synthesis in unbounded scenes,'' \emph{{ACM} Trans. Graph.}, vol.~42, no.~4, pp. 89:1--89:12, 2023.

\bibitem{Blumenthal-BarbyE14}
D.~Blumenthal{-}Barby and P.~Eisert, ``High-resolution depth for binocular image-based modeling,'' \emph{Comput. Graph.}, vol.~39, pp. 89--100, 2014.

\bibitem{CampbellVHC08}
N.~D.~F. Campbell, G.~Vogiatzis, C.~Hern{\'{a}}ndez, and R.~Cipolla, ``Using multiple hypotheses to improve depth-maps for multi-view stereo,'' in \emph{{ECCV}}, D.~A. Forsyth, P.~H.~S. Torr, and A.~Zisserman, Eds., 2008, pp. 766--779.

\bibitem{GallianiLS15}
S.~Galliani, K.~Lasinger, and K.~Schindler, ``Massively parallel multiview stereopsis by surface normal diffusion,'' in \emph{{ICCV}}, 2015, pp. 873--881.

\bibitem{TolaSF12}
E.~Tola, C.~Strecha, and P.~Fua, ``Efficient large-scale multi-view stereo for ultra high-resolution image sets,'' \emph{Mach. Vis. Appl.}, vol.~23, no.~5, pp. 903--920, 2012.

\bibitem{VuLPK12}
H.~Vu, P.~Labatut, J.~Pons, and R.~Keriven, ``High accuracy and visibility-consistent dense multiview stereo,'' \emph{{IEEE} Trans. Pattern Anal. Mach. Intell.}, vol.~34, no.~5, pp. 889--901, 2012.

\bibitem{ChenLGSLJF19}
W.~Chen, H.~Ling, J.~Gao, E.~J. Smith, J.~Lehtinen, A.~Jacobson, and S.~Fidler, ``Learning to predict 3d objects with an interpolation-based differentiable renderer,'' in \emph{NeurIPS}, 2019, pp. 9605--9616.

\bibitem{Liu0LL19}
S.~Liu, W.~Chen, T.~Li, and H.~Li, ``Soft rasterizer: {A} differentiable renderer for image-based 3d reasoning,'' in \emph{{ICCV}}, 2019, pp. 7707--7716.

\bibitem{EstebanS03}
C.~H. Esteban and F.~Schmitt, ``Silhouette and stereo fusion for 3d object modeling,'' in \emph{{3DIM}}, 2003, pp. 46--53.

\bibitem{FuaL95}
P.~Fua and Y.~G. Leclerc, ``Object-centered surface reconstruction: Combining multi-image stereo and shading,'' \emph{Int. J. Comput. Vis.}, vol.~16, no.~1, pp. 35--56, 1995.

\bibitem{LiaoDG18}
Y.~Liao, S.~Donn{\'{e}}, and A.~Geiger, ``Deep marching cubes: Learning explicit surface representations,'' in \emph{{CVPR}}, 2018, pp. 2916--2925.

\bibitem{MunkbergCHES0GF22}
J.~Munkberg, W.~Chen, J.~Hasselgren, A.~Evans, T.~Shen, T.~M{\"{u}}ller, J.~Gao, and S.~Fidler, ``Extracting triangular 3d models, materials, and lighting from images,'' in \emph{{CVPR}}, 2022, pp. 8270--8280.

\bibitem{ShenGYLF21}
T.~Shen, J.~Gao, K.~Yin, M.~Liu, and S.~Fidler, ``Deep marching tetrahedra: a hybrid representation for high-resolution 3d shape synthesis,'' in \emph{NeurIPS}, 2021, pp. 6087--6101.

\bibitem{DBLP:conf/cvpr/JiangSMHNF20}
C.~M. Jiang, A.~Sud, A.~Makadia, J.~Huang, M.~Nie{\ss}ner, and T.~A. Funkhouser, ``Local implicit grid representations for 3d scenes,'' in \emph{CVPR}, 2020, pp. 6000--6009.

\bibitem{LinWRSKFL19}
C.~Lin, O.~Wang, B.~C. Russell, E.~Shechtman, V.~G. Kim, M.~Fisher, and S.~Lucey, ``Photometric mesh optimization for video-aligned 3d object reconstruction,'' in \emph{{CVPR}}, 2019, pp. 969--978.

\bibitem{WenZCLXF23}
C.~Wen, Y.~Zhang, C.~Cao, Z.~Li, X.~Xue, and Y.~Fu, ``Pixel2mesh++: 3d mesh generation and refinement from multi-view images,'' \emph{{IEEE} Trans. Pattern Anal. Mach. Intell.}, vol.~45, no.~2, pp. 2166--2180, 2023.

\bibitem{WenZLF19}
C.~Wen, Y.~Zhang, Z.~Li, and Y.~Fu, ``Pixel2mesh++: Multi-view 3d mesh generation via deformation,'' in \emph{{ICCV}}, 2019, pp. 1042--1051.

\bibitem{mipnerf}
J.~T. Barron, B.~Mildenhall, M.~Tancik, P.~Hedman, R.~Martin{-}Brualla, and P.~P. Srinivasan, ``Mip-nerf: {A} multiscale representation for anti-aliasing neural radiance fields,'' in \emph{ICCV}, 2021, pp. 5835--5844.

\bibitem{DBLP:conf/cvpr/BarronMVSH22}
J.~T. Barron, B.~Mildenhall, D.~Verbin, P.~P. Srinivasan, and P.~Hedman, ``Mip-nerf 360: Unbounded anti-aliased neural radiance fields,'' in \emph{CVPR}, 2022, pp. 5460--5469.

\bibitem{DBLP:conf/iccv/ReiserPL021}
C.~Reiser, S.~Peng, Y.~Liao, and A.~Geiger, ``Kilonerf: Speeding up neural radiance fields with thousands of tiny mlps,'' in \emph{ICCV}, 2021, pp. 14\,315--14\,325.

\bibitem{DBLP:conf/iccv/GarbinK0SV21}
S.~J. Garbin, M.~Kowalski, M.~Johnson, J.~Shotton, and J.~P.~C. Valentin, ``Fastnerf: High-fidelity neural rendering at 200fps,'' in \emph{ICCV}, 2021, pp. 14\,326--14\,335.

\bibitem{DBLP:journals/tog/MullerESK22}
T.~M{\"{u}}ller, A.~Evans, C.~Schied, and A.~Keller, ``Instant neural graphics primitives with a multiresolution hash encoding,'' \emph{{ACM} Trans. Graph.}, vol.~41, no.~4, pp. 102:1--102:15, 2022.

\bibitem{DBLP:conf/eccv/ChenXGYS22}
A.~Chen, Z.~Xu, A.~Geiger, J.~Yu, and H.~Su, ``Tensorf: Tensorial radiance fields,'' in \emph{ECCV}, 2022, pp. 333--350.

\bibitem{DBLP:conf/cvpr/Fridovich-KeilM23}
S.~Fridovich{-}Keil, G.~Meanti, F.~R. Warburg, B.~Recht, and A.~Kanazawa, ``K-planes: Explicit radiance fields in space, time, and appearance,'' in \emph{CVPR}, 2023, pp. 12\,479--12\,488.

\bibitem{r3dgs}
J.~Gao, C.~Gu, Y.~Lin, H.~Zhu, X.~Cao, L.~Zhang, and Y.~Yao, ``Relightable 3d gaussian: Real-time point cloud relighting with {BRDF} decomposition and ray tracing,'' in \emph{ECCV}, 2024.

\bibitem{MeschederONNG19}
L.~M. Mescheder, M.~Oechsle, M.~Niemeyer, S.~Nowozin, and A.~Geiger, ``Occupancy networks: Learning 3d reconstruction in function space,'' in \emph{{CVPR}}, 2019, pp. 4460--4470.

\bibitem{OechsleP021}
M.~Oechsle, S.~Peng, and A.~Geiger, ``{UNISURF:} unifying neural implicit surfaces and radiance fields for multi-view reconstruction,'' in \emph{{ICCV}}, 2021, pp. 5569--5579.

\bibitem{DBLP:conf/eccv/PengNMP020}
S.~Peng, M.~Niemeyer, L.~M. Mescheder, M.~Pollefeys, and A.~Geiger, ``Convolutional occupancy networks,'' in \emph{ECCV}, 2020, pp. 523--540.

\bibitem{ViciniSJ22}
D.~Vicini, S.~Speierer, and W.~Jakob, ``Differentiable signed distance function rendering,'' \emph{{ACM} Trans. Graph.}, vol.~41, no.~4, pp. 125:1--125:18, 2022.

\bibitem{neus}
P.~Wang, L.~Liu, Y.~Liu, C.~Theobalt, T.~Komura, and W.~Wang, ``Neus: Learning neural implicit surfaces by volume rendering for multi-view reconstruction,'' in \emph{NeurIPS}, 2021, pp. 27\,171--27\,183.

\bibitem{YarivKMGABL20}
L.~Yariv, Y.~Kasten, D.~Moran, M.~Galun, M.~Atzmon, R.~Basri, and Y.~Lipman, ``Multiview neural surface reconstruction by disentangling geometry and appearance,'' in \emph{NeurIPS}, 2020.

\bibitem{DBLP:conf/nips/Fu0OT22}
Q.~Fu, Q.~Xu, Y.~S. Ong, and W.~Tao, ``Geo-neus: Geometry-consistent neural implicit surfaces learning for multi-view reconstruction,'' in \emph{NeurIPS}, 2022.

\bibitem{DBLP:conf/iclr/WuWPXTLL23}
T.~Wu, J.~Wang, X.~Pan, X.~Xu, C.~Theobalt, Z.~Liu, and D.~Lin, ``Voxurf: Voxel-based efficient and accurate neural surface reconstruction,'' in \emph{ICLR}, 2023.

\bibitem{DBLP:conf/cvpr/Li0ETU0L23}
Z.~Li, T.~M{\"{u}}ller, A.~Evans, R.~H. Taylor, M.~Unberath, M.~Liu, and C.~Lin, ``Neuralangelo: High-fidelity neural surface reconstruction,'' in \emph{CVPR}, 2023, pp. 8456--8465.

\bibitem{DBLP:conf/iccv/YuLT0NK21}
A.~Yu, R.~Li, M.~Tancik, H.~Li, R.~Ng, and A.~Kanazawa, ``Plenoctrees for real-time rendering of neural radiance fields,'' in \emph{ICCV}, 2021, pp. 5732--5741.

\bibitem{DBLP:conf/cvpr/Fridovich-KeilY22}
S.~Fridovich{-}Keil, A.~Yu, M.~Tancik, Q.~Chen, B.~Recht, and A.~Kanazawa, ``Plenoxels: Radiance fields without neural networks,'' in \emph{CVPR}, 2022, pp. 5491--5500.

\bibitem{DBLP:conf/nips/YarivGKL21}
L.~Yariv, J.~Gu, Y.~Kasten, and Y.~Lipman, ``Volume rendering of neural implicit surfaces,'' in \emph{NeurIPS}, 2021, pp. 4805--4815.

\bibitem{DBLP:conf/nips/YuPNS022}
Z.~Yu, S.~Peng, M.~Niemeyer, T.~Sattler, and A.~Geiger, ``Monosdf: Exploring monocular geometric cues for neural implicit surface reconstruction,'' in \emph{NeurIPS}, 2022.

\bibitem{neudf}
Y.~Liu, L.~Wang, J.~Yang, W.~Chen, X.~Meng, B.~Yang, and L.~Gao, ``Neudf: Leaning neural unsigned distance fields with volume rendering,'' in \emph{CVPR}, 2023, pp. 237--247.

\bibitem{d1nerf}
K.~Deng, A.~Liu, J.~Zhu, and D.~Ramanan, ``Depth-supervised nerf: Fewer views and faster training for free,'' in \emph{CVPR}, 2022, pp. 12\,872--12\,881.

\bibitem{d2nerf}
B.~Roessle, J.~T. Barron, B.~Mildenhall, P.~P. Srinivasan, and M.~Nie{\ss}ner, ``Dense depth priors for neural radiance fields from sparse input views,'' in \emph{CVPR}, 2022, pp. 12\,882--12\,891.

\bibitem{nerfingmvs}
Y.~Wei, S.~Liu, Y.~Rao, W.~Zhao, J.~Lu, and J.~Zhou, ``Nerfingmvs: Guided optimization of neural radiance fields for indoor multi-view stereo,'' in \emph{ICCV}, 2021, pp. 5590--5599.

\bibitem{DBLP:journals/tog/ThiesZN19}
J.~Thies, M.~Zollh{\"{o}}fer, and M.~Nie{\ss}ner, ``Deferred neural rendering: image synthesis using neural textures,'' \emph{{ACM} Trans. Graph.}, vol.~38, no.~4, pp. 66:1--66:12, 2019.

\bibitem{Wang_2021_CVPR}
Q.~Wang, Z.~Wang, K.~Genova, P.~P. Srinivasan, H.~Zhou, J.~T. Barron, R.~Martin-Brualla, N.~Snavely, and T.~Funkhouser, ``Ibrnet: Learning multi-view image-based rendering,'' in \emph{CVPR}, 2021, pp. 4690--4699.

\bibitem{DBLP:conf/iclr/ThiesZTSN20}
J.~Thies, M.~Zollh{\"{o}}fer, C.~Theobalt, M.~Stamminger, and M.~Nie{\ss}ner, ``Image-guided neural object rendering,'' in \emph{ICLR}, 2020.

\bibitem{nerftex}
Y.~Huang, Y.~Cao, Y.~Lai, Y.~Shan, and L.~Gao, ``Nerf-texture: Texture synthesis with neural radiance fields,'' in \emph{{SIGGRAPH}}, 2023, pp. 43:1--43:10.

\bibitem{rakotosaona2023nerfmeshing}
M.-J. Rakotosaona, F.~Manhardt, D.~M. Arroyo, M.~Niemeyer, A.~Kundu, and F.~Tombari, ``Nerfmeshing: Distilling neural radiance fields into geometrically-accurate 3d meshes,'' in \emph{International Conference on 3D Vision (3DV)}, 2023.

\bibitem{openmvs}
\BIBentryALTinterwordspacing
D.~Cernea, ``{OpenMVS}: Multi-view stereo reconstruction library,'' 2020. [Online]. Available: \url{https://github.com/cdcseacave/openMVS}
\BIBentrySTDinterwordspacing

\bibitem{nvdiffrast}
S.~Laine, J.~Hellsten, T.~Karras, Y.~Seol, J.~Lehtinen, and T.~Aila, ``Modular primitives for high-performance differentiable rendering,'' \emph{{ACM} Trans. Graph.}, vol.~39, no.~6, pp. 194:1--194:14, 2020.

\bibitem{schoenberger2016mvs}
J.~L. Sch\"{o}nberger, E.~Zheng, M.~Pollefeys, and J.-M. Frahm, ``Pixelwise view selection for unstructured multi-view stereo,'' in \emph{ECCV}, 2016.

\bibitem{schoenberger2016sfm}
J.~L. Sch\"{o}nberger and J.-M. Frahm, ``Structure-from-motion revisited,'' in \emph{CVPR}, 2016.

\bibitem{flip}
P.~Andersson, J.~Nilsson, T.~Akenine{-}M{\"{o}}ller, M.~Oskarsson, K.~{\AA}str{\"{o}}m, and M.~D. Fairchild, ``{FLIP:} {A} difference evaluator for alternating images,'' \emph{Proc. {ACM} Comput. Graph. Interact. Tech.}, vol.~3, no.~2, pp. 15:1--15:23, 2020.

\end{thebibliography}

\end{document}